%% file: paper.tex
\documentclass{article}

\PassOptionsToPackage{numbers,sort&compress}{natbib}
\usepackage{dualverse}
\usepackage[utf8]{inputenc}
\usepackage[T1]{fontenc}
\usepackage{lmodern}

\usepackage{url}
\usepackage{amsmath,amsfonts,amssymb,amsthm,bm}
\usepackage{nicefrac}
\usepackage{microtype}
\usepackage{graphicx}
\usepackage{float}
\usepackage{longtable,array,multirow}
\usepackage{subcaption}
\usepackage{placeins}
\usepackage{enumitem}
\usepackage{xcolor}
\floatstyle{ruled}
\newfloat{algorithm}{tbp}{loa}
\floatname{algorithm}{Algorithm}

\newcommand{\CL}{CopyLoop}
\newcommand{\LP}{LoopSpec}
\newcommand{\cat}{\mathbin{\|}}

\newcommand{\proc}[1]{\textsc{#1}}

\usepackage[capitalize,noabbrev]{cleveref}

\title{CopyLoop: Fast Looped Transformer Decoding via Context Copying and Shallow Screening}
\SetHeaderTitle{CopyLoop: Fast Looped Transformer Decoding via Context Copying and Shallow Screening}
\SetPaperDate{28-Oct-2026}
\SetCorrespondence{info@dualverse.ai}
\SetPaperLogo{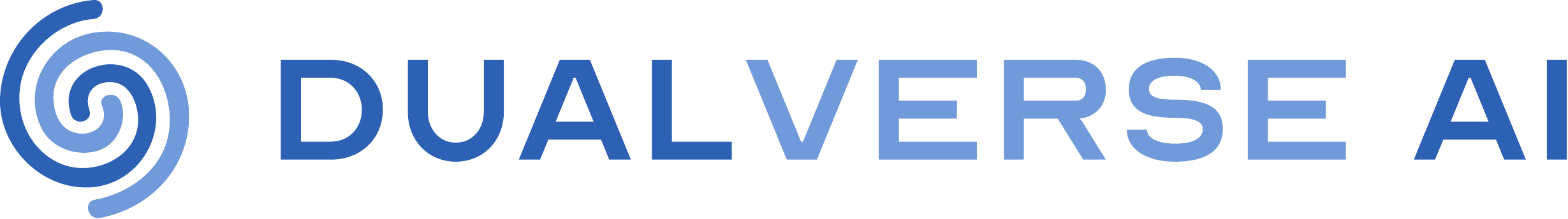}
\SetPDFAuthor{Wenyu Du, Stephen Chung}
\SetDualverseAuthors{
  Wenyu Du \\DualverseAI
  \and Stephen Chung\\DualverseAI
}{
  {\sffamily\bfseries\color{DualverseA}
    Wenyu Du \quad\quad Stephen Chung
  }\\[1em]
  \parbox{0.9\linewidth}{\centering
    \normalfont\sffamily\bfseries\small\color{DualverseB}
    DualverseAI
  }
}

\begin{document}

\input{sections/abstract}
\maketitle
\begingroup
\renewcommand{\thefootnote}{}
\footnotetext[1]{This work extends a human-verified finding made by Station (\mbox{\url{https://dualverse.ai/station}}) as part of our \href{https://dualverse.ai/auto_research}{Autonomous Research Program} (\mbox{\url{https://dualverse.ai/auto_research}}). See Section~\ref{sec:ai_use} for a discussion of AI use.}
\endgroup

\vspace{1.8ex}
\input{sections/introduction}
\input{sections/method}
\input{sections/results}
\input{sections/related}

\input{sections/conclusion}
\input{sections/ai_use}

\clearpage
\bibliographystyle{unsrtnat}
\bibliography{references}

\newpage
\appendix
\renewcommand{\theHtable}{A\arabic{table}}
\renewcommand{\theHfigure}{A\arabic{figure}}

\input{sections/appendix}

\end{document}

%% file: sections/abstract.tex
\begin{abstract}
Looped Transformers increase computational capacity through the repeated application of shared blocks, but their recurrent depth makes conventional full-depth decoding expensive. 
While existing speculative pipelines accelerate generation by processing consecutive token positions at varying depths, they are typically limited by sequential decoding overheads. 
We introduce \textbf{CopyLoop}, a training-free decoding algorithm that combines low-cost context-copy proposals with shallow screening and continuous recurrent scheduling. 
Rather than step-by-step generation, CopyLoop proposes multi-token suffixes from repeated context. It then evaluates them early at a shallow recurrent depth to efficiently filter out incorrect branches, processing newer blocks concurrently while earlier blocks are still progressing toward full-depth verification. 
We evaluate CopyLoop across six model checkpoints from the Ouro, Raven, and Huginn families, covering code generation and multi-step mathematical reasoning. 
CopyLoop achieves decoding speedups of $2.98\times$--$10.70\times$ over full-depth autoregressive decoding on Base models and $2.74\times$--$3.52\times$ on Thinking models, consistently outperforming or closely matching all competing baselines across evaluation tasks. 
Furthermore, ablation studies confirm a strong synergy between shallow screening and continuous scheduling, while our analyses show that CopyLoop requires only lightweight model calibration and yields greater speedups on longer generations with reusable contexts.
\end{abstract}

%% file: sections/introduction.tex
\section{Introduction}

Looped Transformers increase computational capacity without scaling parameter count by repeatedly applying shared Transformer blocks~\citep{Universal19,Programmable23}. Recent studies demonstrate that they can outperform conventional Transformers of equivalent size on specific tasks, highlighting the potential of recurrent computation to elevate model performance~\citep{Huginn25,Ouro25,Raven25}. Nevertheless, this efficiency comes at a price: every token needs complete all recurrent passes before the next can initiate, causing decoding latency to scale with recurrent depth.

Intermediate predictions circumvent this bottleneck by decoding hidden states from early recurrent passes into draft tokens, allowing them to start their own processing cycles while current positions proceed deeper. Self-speculative decoding frameworks~\citep{LoopSpec26,WaveFront26} exploit this by concurrently processing consecutive token positions at varying depths. Once a position attains full depth, its final next-token prediction verifies the draft: a match preserves the continuation, whereas a mismatch discards it. When intermediate predictions prove accurate, such pipelining drastically trims sequential computation.

The success of prior self-speculative methods relies on intermediate recurrent passes to generate draft tokens step by step, which remains limited by sequential decoding overheads. 
We broaden this perspective: rather than generating drafts token by token, we observe that direct copy-paste from repeated context—such as recurrent phrases or structured blocks—provides a much cheaper and richer source of multi-token proposals, completely bypassing stepwise drafting. 
Because such copied guesses can be wrong, we introduce shallow recurrent verification to cheaply filter out invalid suffixes before full-depth processing. 
By coupling these low-cost copy-paste proposals with an efficient shallow screening schedule, we call this algorithm \textit{\CL}. 

\begin{figure}[t]
\centering
\includegraphics[width=.9\linewidth]{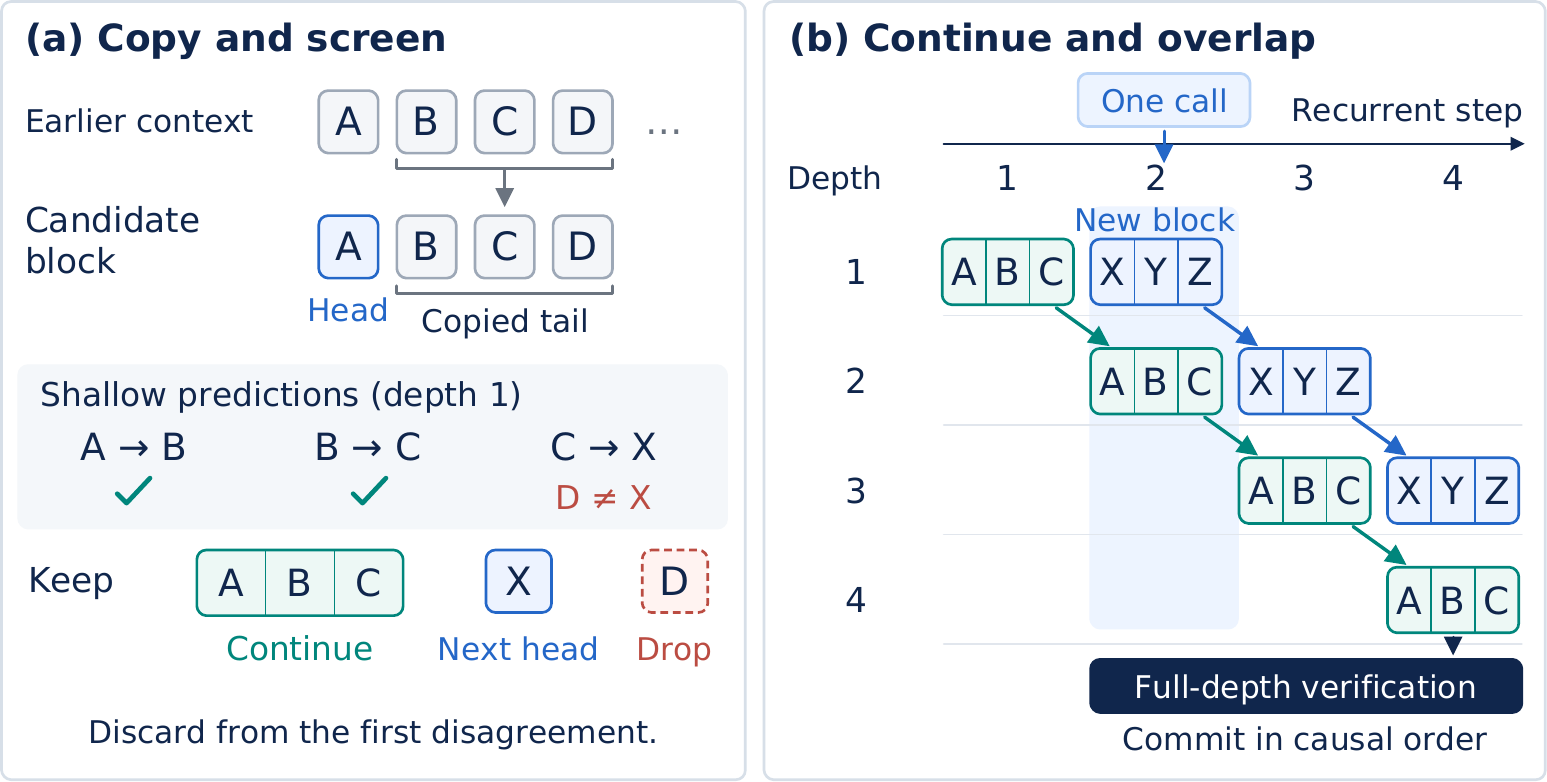}
\caption{\textbf{Screen copied tokens while full-depth verification continues.} (a) A shallow-predicted token A is extended by B C D from earlier context. Predecessor predictions A$\to$B, B$\to$C, and C$\to$X retain A B C, discard D, and supply X as the next head. (b) Retained states advance alongside the next block. Each column represents one pass of the shared recurrent block. Full-depth predecessor predictions commit final tokens in causal order. This teaching example uses $d_s=1$ and $R=4$.}
\label{fig:method}
\end{figure}

Concretely, \CL\ works as follows. At each step, a shallow prediction forms the initial token of a block, while a context-matched suffix supplies the remaining tokens. 
The entire block then advances jointly to an early recurrent depth, where each copied token is screened against its predecessor's next-token prediction: only the longest agreeing prefix is preserved, and the rest is discarded. 
The surviving tokens continue toward full-depth verification, while the retained position immediately seeds the next block, enabling newer proposals to enter the pipeline before earlier blocks complete verification. 
In this manner, \CL\ advances multiple positions per readout, pruning unreliable suffixes before they trigger expensive deeper computation. 
\CL\ uses existing model readouts and requires no additional training. 
Figure~\ref{fig:method} illustrates this mechanism through a four-token example. 

We evaluate \CL\ across six model checkpoints from the Ouro, Raven, and Huginn families, covering different recurrent depths, code generation, and mathematical reasoning with both Base and Thinking models. 
Across our evaluations, \CL\ achieves decoding speedups of $2.98\text{--}10.70\times$ over full-depth autoregressive decoding on Base models and $2.74\text{--}3.52\times$ on Thinking models, outperforming all baselines in 14 of 16 Base model--task configurations and all eight Thinking configurations, and closely matching them on the remaining configurations. 
Beyond these overall gains, ablation studies confirm a strong synergy between shallow screening and continuous scheduling: screening yields substantially larger latency reductions when new blocks enter before full-depth verification. 
Furthermore, hyperparameter studies demonstrate that \CL\ requires only lightweight calibration when transferring to new model architectures. 
Lastly, our analysis shows that \CL\ delivers greater speedups on longer lengths and reusable spans, whereas short-response tasks see limited acceleration due to pipeline overheads.\footnote{Code is available at \url{https://github.com/dualverse-ai/copyloop}.}

Our contributions are:
\begin{itemize}
    \item {Copy-paste speculation with shallow screening.}
    We introduce \CL, a training-free efficient decoding algorithm for looped Transformers that combines context-derived multi-token proposals with shallow screening and continuous recurrent scheduling, using causal full-depth verification to output final tokens.

    \item {Evaluation across architectures and generation modes.}
    We evaluate decoding efficiency and task accuracy across four Base and two Thinking checkpoints, demonstrating significant acceleration over autoregressive decoding and gains over existing self-speculative pipelines across code and reasoning workloads.

\item {Analysis of mechanism and hyperparameter stability.} 
We reveal the synergy between shallow screening and continuous scheduling via ablation studies, verify the stability of core hyperparameters, and quantify how reusable span patterns and response lengths shape overall speedups.
\end{itemize}

%% file: sections/method.tex
\section{Method}
\label{sec:method}

\paragraph{Block Construction.} 
To bypass the overhead of stepwise drafting, \CL\ constructs multi-token candidate blocks by combining fresh model predictions with reusable historical context. 
Specifically, to form a candidate block, \CL\ pairs a model-predicted \emph{head} with a context-copied \emph{tail}. 
Given a sequence prefix $p$ whose last position predicts a head token $h$:
\begin{enumerate}
    \item {Suffix Matching:} Following prompt lookup decoding~\citep{PLD23}, we search the combined sequence $p \mathbin{\Vert} h$ for its longest matching suffix in the preceding context (evaluating lengths from $4$ down to $1$).
    \item {Tail Copying:} We copy up to $K$ subsequent tokens from that matching location as a contiguous tail.
    \item {Fallback:} This yields a candidate block $B = (b_0, b_1, \ldots, b_k)$ with $b_0 = h$ and $0 \leq k \leq K$. If no matching suffix is found, the block gracefully falls back to a single-token block $B = (h)$.
\end{enumerate}

\paragraph{Shallow Screening.} 
To minimize unnecessary compute on incorrect context-copied tails, \CL\ introduces a fast screening mechanism before full-depth processing. 
In a Looped Transformer with $D$ total recurrent passes, all tokens in the candidate block $B$ advance together to the screening depth $d_s < D$ under causal attention over $p \mathbin{\Vert} B$. 
Let $g_d(\cdot)$ denote the greedy next-token prediction at recurrent depth $d$. 
At depth $d_s$, we screen each copied token against the shallow prediction of its immediate predecessor through the following steps:
\begin{enumerate}
    \item {Valid-Token Screening:} We identify the longest continuous span starting from the head where each copied token matches its predecessor's shallow prediction:
    \begin{equation}
    m =
    \max \left\{
    r \in \{0,\ldots,k\}
    \;\middle|\;
    b_j =
    g_{d_s}\!\left(
    p \mathbin{\Vert} b_0
    \mathbin{\Vert} \cdots
    \mathbin{\Vert} b_{j-1}
    \right)
    \text{ for all } 1\leq j\leq r
    \right\}.
    \label{eq:screen}
    \end{equation}
    \item {Survival and Pruning:} The head $b_0$ and the first $m$ tokens that pass screening survive and continue to deeper layers. The remaining suffix $(b_{m+1},\ldots,b_k)$ is pruned, and its KV and hidden states are immediately released.
    \item {Seeding the Next Block:} Regardless of tail truncation, the shallow prediction at the final surviving position, $g_{d_s}(p \mathbin{\Vert} b_0 \mathbin{\Vert}\cdots\mathbin{\Vert} b_m)$, is repurposed as the head of the next block.
\end{enumerate}
As illustrated in Figure~\ref{fig:method} for block $\text{A B C D}$, when shallow readouts at $\text{A}$, $\text{B}$, and $\text{C}$ yield $\text{B}$, $\text{C}$, and $\text{X}$, positions $\text{B}$ and $\text{C}$ match their predecessors' predictions while $\text{D}$ conflicts with $\text{X}$. Consequently, \CL\ retains $\text{A B C}$, discards $\text{D}$, and uses $\text{X}$ as the next block's head.

\paragraph{Concurrent Pipeline.} 
While surviving positions preserve their depth-$d_s$ states and continue toward full depth $D$, new candidate blocks continuously enter the pipeline.  
At each recurrent step, the shared Transformer block advances all active yet incomplete positions by one pass, processing earlier and incoming blocks concurrently. 
Because each position accesses its own specific KV states, this setup naturally reuses existing caches and overlaps shallow screening with deep verification without breaking causal dependencies. 
Finally, positions reaching depth $D$ retain their final readouts, awaiting full-depth verification.

\paragraph{Full-Depth Verification.} 
At depth $D$, full-depth verification applies the same checking logic as shallow screening while outputting final tokens. 
Starting from the verified prefix, matching candidate tokens are accepted sequentially. 
At the first mismatch, the full-depth prediction supplies the correct token, discarding the invalid suffix along with all dependent continuations. 
Conversely, if the entire copied tail matches, the final position's prediction yields one additional accepted token. 
Finally, decoding proceeds from any cached continuation consistent with the verified prefix; if none is available, a new block starts using the last accepted token as its head. 
Appendix~\ref{app:procedure} provides the pseudocode and a straightforward extension that branches at a second readout depth.

%% file: sections/results.tex
\section{Experiments}
\label{sec:experiments}
We evaluate \CL's decoding efficiency across diverse looped Transformer architectures, spanning code generation and mathematical reasoning with both Base and Thinking models. 
Controlled ablations isolate the synergy between shallow screening and continuous scheduling, while hyperparameter studies demonstrate that \CL\ requires only lightweight model-level calibration. 
We further analyze how repetition structure and response length influence speedup gains, identifying specific conditions where pipeline overheads constrain acceleration. Appendices~\ref{app:protocol} and~\ref{app:performance} provide experimental protocols and supplementary results.

\paragraph{Models, tasks, and baselines.}
We evaluate our method across six looped Transformer checkpoints, spanning both standard generation and explicit multi-step reasoning:

\begin{itemize}[leftmargin=*, itemsep=2pt, topsep=2pt]
    \item Models: We examine four non-thinking checkpoints: Ouro-1.4B and Ouro-2.6B ($R=4$) \cite{Ouro25}, alongside Raven-TinyLlama and Huginn-3.5B ($R=32$) \cite{Raven25,Huginn25} for cross-architecture validation. To assess performance on
    extended reasoning traces, we additionally evaluate the reasoning-tuned Ouro-1.4B-Thinking and Ouro-2.6B-Thinking checkpoints ($R=4$) \cite{Ouro25}.
    All methods share BF16 precision, prompts and stopping rules. 

    \item Tasks: For standard models, our benchmark suite covers code generation via HumanEval+ and MBPP+ (using extended EvalPlus tests) \cite{HumanEval21,MBPP21,EvalPlus23}, and mathematical reasoning via GSM8K \cite{GSM8K21} and
    MATH-500 \cite{MATH21,Verify24}.
    For Thinking models, we curate a challenging reasoning panel combining AIME2024, AIME2025, BeyondAIME, and MATH-500 \cite{MATH21,Verify24} (totaling 188 problems).

    \item Baselines: We benchmark against four reference paradigms:
    (1) AR, standard full-depth autoregressive decoding;
    (2) Prompt Lookup Decoding (PLD) \cite{PLD23}, which serves as a context-matching lookup reference;
    (3) LoopSpec \cite{LoopSpec26}, which proposes candidates at intermediate depths and verifies them at full depth; and
    (4) WaveFront \cite{WaveFront26}, which overlaps drafting and verification via a continuous pipeline. 
\end{itemize}

\paragraph{Experimental Setup.}
All evaluations run under on RTX 5090 GPUs with a concurrency of one request per worker.
Unless specified otherwise, CopyLoop uses a copied-tail limit of $K=6$ for Base models ($K=2$ for Thinking models), with screening depths $d_s=1$ (Ouro), $d_s=4$ (Raven-TinyLlama), and $d_s=8$ (Huginn). 
Section~\ref{sec:hyperparameters} details the calibration and sensitivity analysis. 
Output budgets are set to 1,024 tokens for Base models—except for Math-500, which uses 2,048 tokens to accommodate longer reasoning trajectories—and 8,192 tokens for Thinking models, matched identically across all decoding baselines per task. 
For baseline configurations, we adopt their original settings where applicable (e.g., LoopSpec) or re-tune them for optimal performance on our datasets (e.g., WaveFront and PLD).

We report decoding throughput speedup relative to each model’s native AR baseline. Specifically, decoding throughput is measured as the total generated tokens divided by the total decoding time across all evaluated requests, and throughput speedup is defined as the ratio of our decoding throughput to that of the native AR baseline: 

\begin{equation}
    \text{Speedup}
    =
    \frac{
        \sum_i N_{\mathrm{method},i}
        /
        \sum_i T^{\mathrm{dec}}_{\mathrm{method},i}
    }{
        \sum_i N_{\mathrm{AR},i}
        /
        \sum_i T^{\mathrm{dec}}_{\mathrm{AR},i}
    },
    \label{eq:speedup}
\end{equation}
where $N_i$ denotes the number of generated tokens excluding the first token, and $T^{\mathrm{dec}}_i$ denotes the corresponding decoding time for request $i$.

\subsection{Decoding Performance Across Architectures and Tasks}
\label{sec:overall}

Table~\ref{tab:overall} compares AR, PLD, LoopSpec, WaveFront and CopyLoop across four Base models and two Thinking checkpoints, reporting both task accuracy and decoding speedup. 
Overall, CopyLoop achieves substantial acceleration while preserving competitive task performance across diverse architectures, tasks and reasoning modes.

\input{sections/tables/overall}

\paragraph{Base models.}
CopyLoop delivers consistent speedups ranging from $2.98\times$ to $10.70\times$ over AR, outperforming all three baselines in 14 of the 16 evaluated model--task configurations. 
On Ouro models, the most significant gains appear in code generation on HumanEval+, reaching $6.04\times$ (Ouro-1.4B) and $6.34\times$ (Ouro-2.6B)—translating to roughly $1.86\times$ the throughput of existing speculative baselines. 
The advantage is more modest on MBPP+ (e.g., matching or slightly trailing existing baselines), pointing to the influence of response length and context structure, as analyzed in Section~\ref{sec:regimes}.

The acceleration successfully generalizes to deeper recurrent architectures ($R=32$). 
Specifically, Raven-TinyLlama achieves speedups of $6.71\times$–$10.70\times$, and Huginn-3.5B achieves $3.10\times$–$5.98\times$ over AR, surpassing the faster of the two reference baselines by up to $63.4\%$ and $45.9\%$ respectively. 

\paragraph{Thinking models.}
For explicitly reasoning-tuned checkpoints, CopyLoop accelerates generation by $2.74\times$–$3.23\times$ over AR for Ouro-1.4B-Thinking and $3.14\times$–$3.52\times$ for Ouro-2.6B-Thinking. 
It exceeds all three baselines across all eight reasoning panels, achieving up to approximately $20\%$ higher decoding throughput than the strongest alternative. 
Notable highlights include MATH-500 speedups of $3.23\times$ (1.4B) and $3.52\times$ (2.6B). 
These results demonstrate that CopyLoop's efficiency pipeline extends effectively from standard generation to long-horizon multi-step reasoning traces.

Task accuracies vary across decoding methods. Appendix~\ref{app:numerics} presents diagnostics illustrating how numerical differences under BF16 execution can alter greedy decoding.

\subsection{Ablation Studies on Screening and Scheduling}
\label{sec:interaction}

To isolate the individual and joint contributions of our core components, we conduct a $2\times2$ factorial ablation. 
Specifically, we cross \textbf{shallow screening} (enabled vs.\ disabled) with \textbf{pipeline scheduling} (\textbf{continuous}: new blocks enter before full-depth verification vs.\ \textbf{staged}: admission waits for full verification), while keeping context-copy proposals and state continuation identical across all configurations. 
Our evaluation spans diverse architectures and reasoning tasks, testing Ouro-1.4B and Raven-TinyLlama on HumanEval+ and GSM8K, as well as Ouro-1.4B-Thinking and Ouro-2.6B-Thinking on HumanEval+ and MATH-500. 
Each panel averages 32 problems over two seeds, and we report decode-throughput speedup relative to each model's native AR baseline.

\begin{figure}[t]
\centering
\includegraphics[width=\linewidth]{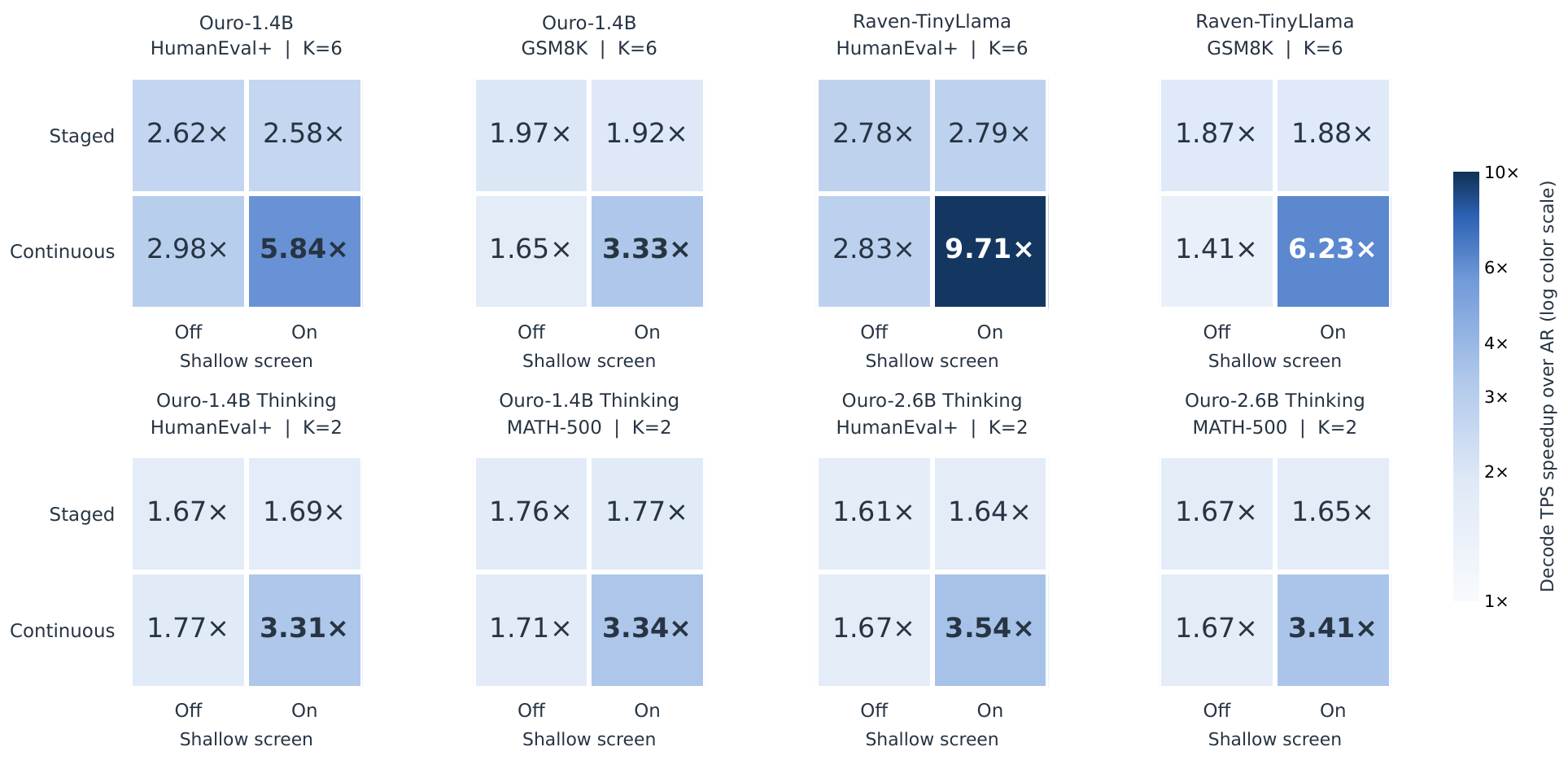}
\caption{\textbf{Shallow screening and continuous scheduling provide complementary gains.} Each cell reports decode-throughput speedup relative to AR for the same model and task ($1.00\times$). Rows vary scheduling and columns vary shallow screening. Bold lower-right cells denote the full method, which achieves the highest throughput in all eight panels. All configurations share a logarithmic color scale.}
\label{fig:synergy}
\end{figure}

As summarized in Figure~\ref{fig:synergy}, the $2\times2$ factorial heatmap reveals a consistent and significant synergy across all eight evaluated panels. 
Under staged scheduling, enabling shallow screening yields negligible performance shifts (ranging from $-2.5\%$ to $+1.9\%$). 
In contrast, when coupled with continuous scheduling, screening unlocks substantial throughput multipliers, boosting performance by $1.87\times$ to $4.42\times$ relative to the corresponding no-screen configuration. 
The reverse dependency is equally clear: while continuous scheduling delivers substantial gains ($1.73\times$ to $3.48\times$) when screening guards the pipeline, its benefits are largely neutralized ($0.75\times$ to $1.14\times$) when screening is disabled. 
Consequently, the fully coupled configuration achieves the highest throughput across all evaluated panels, reaching $3.33\times$ to $9.71\times$ AR speedup on Base models and $3.31\times$ to $3.54\times$ on Thinking models. 

This interdependence highlights a key balance: unconstrained continuous admission risks flooding the pipeline with erroneous suffixes, whereas shallow screening acts as a lightweight gatekeeper to prune them instantly. 
Together, early pruning and continuous admission transform inexpensive copied proposals into substantial decoding speedups. 
Appendix~\ref{app:mechanism} provides the complete numerical results for all eight panels in Figure~\ref{fig:synergy}.

\subsection{Hyperparameter Sensitivity and Transferability}
\label{sec:hyperparameters}

CopyLoop introduces two hyperparameters: the screening depth $d_s$ and the copied-tail limit $K$. 
To understand their sensitivity and determine their optimal values, we conduct a grid search over $K \in \{2, 4, 6, 8\}$ and model-specific screening depths ($d_s \in \{1, 2\}$ for Ouro; $d_s \in \{4, 8, 16\}$ for Raven-TinyLlama and Huginn-3.5B) across two distinct data suites: each model's four main evaluation benchmarks (Section~\ref{sec:overall}), and a diverse set of 20 calibration datasets comprising eight fixed examples each. 
For each model, we select the $(d_s, K)$ configuration that maximizes the equal-weight geometric mean of decoding throughput across these 20 calibration datasets.
We then compare these calibration-derived settings with the benchmark-specific optima on our main tasks. We find that 5 out of 6 calibrated configurations match the optimal settings of the evaluation benchmarks, with the sole exception of Ouro-1.4B Thinking, which incurs a negligible deficit of just 3.9\%.
Further details and full results are provided in Appendix~\ref{app:calibration} (Table~\ref{tab:calibrated-configs},\ref{tab:calibration-task-transfer},\ref{tab:depth-transfer-controlled},\ref{tab:copy-limit-tps}).

\begin{table*}[t]
\centering
\caption{\textbf{Task-wise throughput deficits of model-level optimal configurations.} Deficits (\%, lower is better) compare calibrated $d_s$ and default $K$ with each benchmark's best tested pair. Overall uses the geometric mean of throughput ratios across four benchmarks, with eight fixed examples per benchmark.}
\label{tab:hyperparameter-transfer}

\begin{dualversetable}
\setlength{\tabcolsep}{3pt}
\renewcommand{\arraystretch}{1.04}
\setlength{\DualverseHeadPad}{2.6pt}
\begin{subtable}{\linewidth}
\centering
\phantomcaption
\label{tab:hyperparameter-transfer-base}
\begin{tabular}{>{\raggedright\arraybackslash}p{\dimexpr.24\linewidth-2\tabcolsep\relax}*{2}{>{\centering\arraybackslash}p{\dimexpr.045\linewidth-2\tabcolsep\relax}}*{4}{>{\centering\arraybackslash}p{\dimexpr.14\linewidth-2\tabcolsep\relax}}>{\centering\arraybackslash}p{\dimexpr.11\linewidth-2\tabcolsep\relax}}
\toprule
\DualverseHeadRow
\DualverseHead[l]{Model} & \DualverseHead{$d_s$} & \DualverseHead{$K$} & \DualverseHead{HumanEval+} & \DualverseHead{MBPP+} & \DualverseHead{GSM8K} & \DualverseHead{MATH-500} & \DualverseHead{Overall} \\

\rowcolor{DualverseB!30}
\multicolumn{8}{l}{\textbf{Base models}} \\
Ouro-1.4B & 1 & 6 & 5.96 & 18.49 & 8.46 & 0.00 & \textbf{8.48} \\
Ouro-2.6B & 1 & 6 & 2.90 & 12.64 & 4.56 & 0.27 & \textbf{5.21} \\
Raven-TinyLlama & 4 & 6 & 0.00 & 0.00 & 0.00 & 13.38 & \textbf{3.53} \\
Huginn-3.5B & 8 & 6 & 5.63 & 2.86 & 1.84 & 27.39 & \textbf{10.09} \\
\rowcolor{DualverseB!30}
\multicolumn{8}{l}{\textbf{Additional validation based models}} \\
Raven-Llama-3.2 & 4 & 6 & 0.00 & 0.71 & 4.41 & 3.52 & \textbf{2.18} \\
Raven-OLMo-2 & 4 & 6 & 0.92 & 10.46 & 3.72 & 0.00 & \textbf{3.86} \\
\bottomrule
\end{tabular}
\end{subtable}
\par\medskip
\begin{subtable}{\linewidth}
\centering
\phantomcaption
\label{tab:hyperparameter-transfer-thinking}
\begin{tabular}{>{\raggedright\arraybackslash}p{\dimexpr.24\linewidth-2\tabcolsep\relax}*{2}{>{\centering\arraybackslash}p{\dimexpr.045\linewidth-2\tabcolsep\relax}}*{4}{>{\centering\arraybackslash}p{\dimexpr.14\linewidth-2\tabcolsep\relax}}>{\centering\arraybackslash}p{\dimexpr.11\linewidth-2\tabcolsep\relax}}
\toprule
\DualverseHeadRow
\DualverseHead[l]{Model} & \DualverseHead{$d_s$} & \DualverseHead{$K$} & \DualverseHead{AIME 2024} & \DualverseHead{AIME 2025} & \DualverseHead{BeyondAIME} & \DualverseHead{MATH-500} & \DualverseHead{Overall} \\

\rowcolor{DualverseB!30}
\multicolumn{8}{l}{\textbf{Thinking models}} \\
Ouro-1.4B Thinking & 1 & 2 & 10.93 & 0.00 & 6.46 & 4.63 & \textbf{5.59} \\
Ouro-2.6B Thinking & 1 & 2 & 8.87 & 2.33 & 0.99 & 10.18 & \textbf{5.67} \\
\bottomrule
\end{tabular}
\end{subtable}
\end{dualversetable}
\end{table*}

\paragraph{Cross-Task Stability of Screening Depth vs.\ Task Sensitivity of Copied-Tail Limit.}
The calibration result reveals that the two hyperparameters exhibit different patterns of task and architecture sensitivity. 

First, the screening depth $d_s$ is largely model-dependent and exhibits strong cross-task stability. 
Since $d_s$ determines when to interrupt recurrent execution, its optimal value appears closely related to the model's native recurrent depth $R$. 
For instance, all four Ouro variants (which use a shallow recurrent depth of $R=4$) consistently select $d_s=1$. 
In contrast, Raven-TinyLlama and Huginn-3.5B employ a deeper recurrent structure ($R=32$). Because Raven features fewer layers per recurrent block than Huginn, it requires a larger number of recurrent steps to accumulate sufficient representation depth, which naturally aligns with their calibrated selections of $d_s=8$ for Raven and $d_s=4$ for Huginn, respectively. 
Across tasks, this calibrated $d_s$ is remarkably robust, matching the optimal depth in 20 of 24 model-benchmark configurations. For the remaining few exceptions, the throughput drop remains under $6.81\%$, confirming that a single model-level screening depth generalizes well across diverse workloads. 

By contrast, the optimal copied-tail limit $K$ depends more heavily on the task.
For instance, with $d_s=1$, Ouro-1.4B prefers different limits across benchmarks, choosing $K=8, 2, 2,$ and $6$ for HumanEval+, MBPP+, GSM8K, and MATH-500, respectively.
Even so, further inspection shows that decoding throughput remains remarkably stable across a reasonable range of $K$.
For example, as shown in Table~\ref{tab:copy-limit-tps}, keeping the screening depth fixed, testing any of those four $K$ values results in a throughput drop of no more than $5.76\%$ on Ouro-2.6B MATH-500 and $6.32\%$ on Huginn-3.5B GSM8K, compared to their task-specific configurations.

\paragraph{Practical Takeaway: Model-Level Calibration with Mode-Specific Defaults.}
Our findings reveal a clear split: the screening depth $d_s$ depends primarily on the model architecture, whereas the copied-tail limit $K$ correlates more with the task type—yet remains remarkably stable across a wide range of values.
This insight motivates a simple, robust configuration strategy for deployment: instead of tuning hyperparameters per task, practitioners can calibrate $d_s$ just once at the model level on representative calibration data, while adopting fixed, mode-specific defaults ($K=6$ for Base models and $K=2$ for Thinking models) across all downstream tasks.
To measure the trade-off of this simplification, we compare these unified defaults against the best task-specific configurations.
As summarized in Table~\ref{tab:hyperparameter-transfer}, across the 24 evaluation panels for our six models, task-wise throughput deficits average $6.20\%$ (median $4.60\%$), with overall per-model deficits ranging from $3.53\%$ to $10.09\%$.\footnote{Large task deficits (e.g., up to $27.39\%$ on MATH-500) typically stem from BF16 non-determinism rather than systematic slowdowns: minor trajectory shifts across different $K$ choices can occasionally trigger prolonged, repetitive text loops. While such repetitions can artificially inflate measured \CL\ throughput, they do not reflect true task efficiency.}
These results confirm that our decoupling strategy achieves near-optimal acceleration with minimal tuning overhead.

\paragraph{Validation on Larger Raven Models.}
We further test our model-level calibration strategy on two larger Base models, Raven-Llama-3.2 and Raven-OLMo-2, by calibrating $d_s$ across the 20 datasets while keeping $K=6$ fixed.
Both models select $d_s=4$, which matches the optimal depth across all eight evaluation settings.
Compared to an oracle baseline that independently tunes both $d_s$ and $K$ for each benchmark, the maximum throughput drops are limited to $4.41\%$ for Raven-Llama-3.2 and $10.46\%$ for Raven-OLMo-2.
These results confirm the robustness of combining model-level depth calibration with fixed mode-specific defaults.

\begin{figure}[t]
\centering
\includegraphics[width=.85\linewidth]{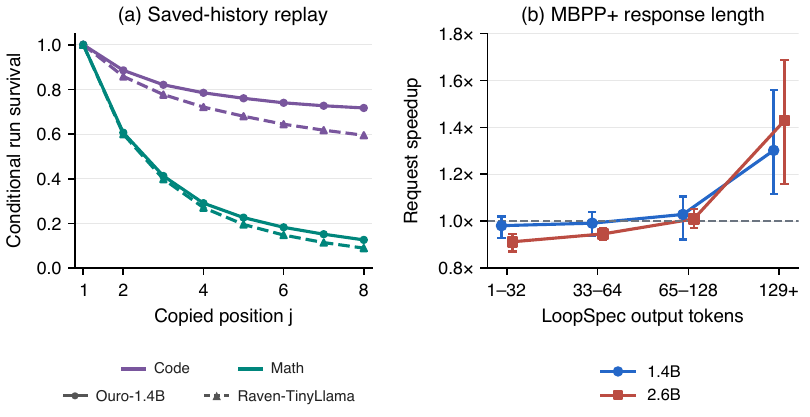}\caption{
\textbf{Request-level gains depend on available continuations and response length.} 
(a) Saved-history replay on HumanEval (code) and GSM8K (math) outputs using a copied-tail limit of $8$. 
Curves show the conditional survival probability $\Pr(M\ge j \mid M\ge 1)$ of matching future tokens across Ouro-1.4B and Raven-TinyLlama. 
(b) MBPP+ request speedup over LoopSpec grouped by output length, with $95\%$ bootstrap confidence intervals (dashed line marks parity). Points are horizontally offset within each length bin for clarity.
}
\label{fig:workloads}
\end{figure}

\subsection{Analysis of Reusable Spans and Response Length}
\label{sec:regimes}

The performance variations across our benchmarks raise a natural question: under what conditions does CopyLoop deliver its best speedups? 
Our analysis reveals that two key factors govern its effectiveness: the availability of reusable, structured text spans and a sufficiently long response to amortize pipeline overheads.

\paragraph{Reusable Span Length.}
We analyze the distribution of matching tokens through static lookups on saved output histories (capped at $8$ tokens). 
As shown in Figure~\ref{fig:workloads}a, code generation (HumanEval) exhibits substantially longer matching spans than mathematical reasoning (GSM8K) across both Ouro-1.4B and Raven-TinyLlama. 
This difference explains why code tasks yield much larger speedups: a single valid copy-paste action covers multiple subsequent positions in parallel, maximizing pipeline throughput. 
Because math tasks feature shorter matching spans, a natural extension of Section~\ref{sec:hyperparameters} is to dynamically adjust the copied-tail limit $K$ using history replay. However, our exploratory experiments show this yields only marginal gains under fixed default budgets. We therefore adopt unified defaults and leave more sophisticated adaptive scheduling for future work.

\paragraph{The Short-Response Boundary.}
However, long matching runs alone do not guarantee lower request latencies. 
On short-response tasks like MBPP+, many generated sequences are simply too brief to amortize the fixed overheads of context lookup, pipeline admission, and startup. 
As illustrated in Figure~\ref{fig:workloads}b, grouping requests by LoopSpec output length reveals a clear boundary: short bins ($\leq 64$ tokens) experience minor slowdowns due to overhead dominance, whereas long bins ($> 128$ tokens) achieve substantial speedups (e.g., $1.43\times$ for Ouro-2.6B). 

Therefore, CopyLoop thrives on workloads characterized by long repetitive patterns and sufficient response length, while encountering natural performance boundaries on tasks dominated by ultra-short answers.

%% file: sections/tables/overall.tex
  \begin{table}[!t]
  \centering
  \caption{\textbf{Accuracy and decoding throughput speedup across architectures and baselines.} Each cell reports accuracy (\%) / speedup relative to AR for the same model and task; AR is normalized to $1.00\times$. Bold values mark the highest speedup at the displayed precision, including ties.}
  \label{tab:overall}
  \begin{dualversetable}
  \setlength{\tabcolsep}{3pt}
  \renewcommand{\arraystretch}{1.04}
  \setlength{\DualverseHeadPad}{2.6pt}
  \begin{subtable}{\linewidth}
  \centering
  \phantomcaption
  \label{tab:overall-base}
  \begin{tabular}{>{\raggedright\arraybackslash}p{\dimexpr.16\linewidth-2\tabcolsep\relax}*{4}{>{\centering\arraybackslash}p{\dimexpr.21\linewidth-2\tabcolsep\relax}}}
  \toprule
  \DualverseHeadRow
  \DualverseHead[l]{Method} & \DualverseHead{HumanEval+} & \DualverseHead{MBPP+} & \DualverseHead{GSM8K} & \DualverseHead{MATH-500} \\

  \rowcolor{DualverseB!30}
  \multicolumn{5}{l}{\textbf{Ouro-1.4B --- Non-thinking}} \\
  AR & 69.51 / 1.00$\times$ & 63.76 / 1.00$\times$ & 78.24 / 1.00$\times$ & 71.60 / 1.00$\times$ \\
  PLD & 69.51 / 2.73$\times$ & 64.02 / 1.40$\times$ & 78.47 / 1.93$\times$ & 72.00 / 1.86$\times$ \\
  \LP & 68.90 / 3.24$\times$ & 63.49 / 2.97$\times$ & 78.77 / 2.95$\times$ & 70.60 / 2.98$\times$ \\
  WaveFront & 68.90 / 3.22$\times$ & 64.02 / 2.84$\times$ & 78.24 / 2.71$\times$ & 72.00 / 2.76$\times$ \\
\textbf{\CL} & 69.51 / \textbf{6.04$\times$} & 63.49 / \textbf{2.98$\times$} & 77.94 / \textbf{3.20$\times$} & 70.60 / \textbf{3.38$\times$} \\

  \rowcolor{DualverseB!30}
  \multicolumn{5}{l}{\textbf{Ouro-2.6B --- Non-thinking}} \\
  AR & 76.22 / 1.00$\times$ & 70.90 / 1.00$\times$ & 86.05 / 1.00$\times$ & 46.60 / 1.00$\times$ \\
  PLD & 74.39 / 2.55$\times$ & 70.90 / 1.38$\times$ & 86.66 / 1.87$\times$ & 47.80 / 1.85$\times$ \\
  \LP & 73.17 / 3.38$\times$ & 71.16 / \textbf{3.11$\times$} & 86.43 / 3.17$\times$ & 43.40 / 3.16$\times$ \\
  WaveFront & 74.39 / 3.40$\times$ & 71.16 / 2.94$\times$ & 86.05 / 3.02$\times$ & 45.20 / 3.03$\times$ \\
\textbf{\CL} & 73.78 / \textbf{6.32$\times$} & 70.63 / 3.08$\times$ & 86.35 / \textbf{3.61$\times$} & 44.00 / \textbf{3.88$\times$} \\

  \rowcolor{DualverseB!30}
  \multicolumn{5}{l}{\textbf{Raven-TinyLlama --- Non-thinking}} \\
  AR & 12.20 / 1.00$\times$ & 25.93 / 1.00$\times$ & 44.50 / 1.00$\times$ & 33.00 / 1.00$\times$ \\
  PLD & 12.20 / 3.22$\times$ & 26.19 / 2.22$\times$ & 45.26 / 2.21$\times$ & 31.60 / 2.34$\times$ \\
  \LP & 12.20 / 6.55$\times$ & 26.46 / 5.99$\times$ & 44.66 / 5.46$\times$ & 32.20 / 6.64$\times$ \\
  WaveFront & 12.20 / 5.44$\times$ & 25.66 / 4.77$\times$ & 44.66 / 3.94$\times$ & 32.40 / 5.12$\times$ \\
\textbf{\CL} & 12.20 / \textbf{10.70$\times$} & 25.93 / \textbf{7.53$\times$} & 44.35 / \textbf{6.71$\times$} & 32.40 / \textbf{8.67$\times$} \\

  \rowcolor{DualverseB!30}
  \multicolumn{5}{l}{\textbf{Huginn-3.5B --- Non-thinking}} \\
  AR & 19.51 / 1.00$\times$ & 30.69 / 1.00$\times$ & 38.82 / 1.00$\times$ & 16.80 / 1.00$\times$ \\
  PLD & 19.51 / 2.05$\times$ & 32.01 / 1.42$\times$ & 39.27 / 1.60$\times$ & 15.80 / 2.69$\times$ \\
  \LP & 20.73 / 3.73$\times$ & 32.54 / \textbf{3.25$\times$} & 40.03 / 3.12$\times$ & 17.00 / 4.10$\times$ \\
  WaveFront & 19.51 / 2.79$\times$ & 32.01 / 2.59$\times$ & 39.27 / 2.56$\times$ & 17.60 / 2.88$\times$ \\
\textbf{\CL} & 19.51 / \textbf{4.67$\times$} & 32.28 / 3.10$\times$ & 39.95 / \textbf{3.37$\times$} & 16.00 / \textbf{5.98$\times$} \\
  \bottomrule
  \end{tabular}
  \end{subtable}
  \par\medskip
  \begin{subtable}{\linewidth}
  \centering
  \phantomcaption
  \label{tab:overall-thinking}
  \begin{tabular}{>{\raggedright\arraybackslash}p{\dimexpr.16\linewidth-2\tabcolsep\relax}*{4}{>{\centering\arraybackslash}p{\dimexpr.21\linewidth-2\tabcolsep\relax}}}
  \toprule
  \DualverseHeadRow
  \DualverseHead[l]{Method} & \DualverseHead{AIME 2024} & \DualverseHead{AIME 2025} & \DualverseHead{BeyondAIME} & \DualverseHead{MATH-500} \\

  \rowcolor{DualverseB!30}
  \multicolumn{5}{l}{\textbf{Ouro-1.4B --- Thinking}} \\
  AR & 20.00 / 1.00$\times$ & 23.33 / 1.00$\times$ & 12.50 / 1.00$\times$ & 81.25 / 1.00$\times$ \\
  PLD & 30.00 / 1.88$\times$ & 20.00 / 1.90$\times$ & 12.50 / 1.96$\times$ & 89.06 / 1.88$\times$ \\
  \LP & 33.33 / 2.64$\times$ & 23.33 / 2.64$\times$ & 15.62 / 2.54$\times$ & 85.94 / 2.79$\times$ \\
  WaveFront & 20.00 / 2.42$\times$ & 26.67 / 2.35$\times$ & 10.94 / 2.34$\times$ & 84.38 / 2.53$\times$ \\
\textbf{\CL} & 23.33 / \textbf{2.91$\times$} & 23.33 / \textbf{3.02$\times$} & 10.94 / \textbf{2.74$\times$} & 82.81 / \textbf{3.23$\times$} \\

  \rowcolor{DualverseB!30}
  \multicolumn{5}{l}{\textbf{Ouro-2.6B --- Thinking}} \\
  AR & 43.33 / 1.00$\times$ & 33.33 / 1.00$\times$ & 14.06 / 1.00$\times$ & 87.50 / 1.00$\times$ \\
  PLD & 46.67 / 1.97$\times$ & 30.00 / 1.93$\times$ & 15.62 / 1.96$\times$ & 89.06 / 1.75$\times$ \\
  \LP & 46.67 / 2.80$\times$ & 23.33 / 2.71$\times$ & 17.19 / 2.67$\times$ & 90.62 / 2.92$\times$ \\
  WaveFront & 43.33 / 2.58$\times$ & 26.67 / 2.50$\times$ & 12.50 / 2.44$\times$ & 85.94 / 2.72$\times$ \\
\textbf{\CL} & 40.00 / \textbf{3.27$\times$} & 36.67 / \textbf{3.14$\times$} & 23.44 / \textbf{3.18$\times$} & 90.62 / \textbf{3.52$\times$} \\
  \bottomrule
  \end{tabular}
  \end{subtable}
  \end{dualversetable}
  \end{table}

%% file: sections/related.tex
\section{Related work}
\label{sec:related}
\paragraph{Recurrent depth and execution.}
Universal Transformers and subsequent looped models reuse parameters across depth~\citep{Universal19,Programmable23,Huginn25,Ouro25,Raven25}. 
Iterative token refinement and attention across recurrent depths offer further opportunities for parallel decoding~\citep{Parallel25,DAS26}. 
Concurrent \LP\ overlaps shallow proposals with full-depth verification and adds a gated second successor~\citep{LoopSpec26}; WaveFront similarly maintains a continuous set of positions at different recurrent depths~\citep{WaveFront26}. 
Concurrent LoopAhead trains a multi-token drafter conditioned on early recurrent states and uses intermediate feedback to suspend and resume candidate branches~\citep{LoopAhead26}. 
In contrast, \CL\ bypasses stepwise drafting overheads entirely, introducing a training-free framework that integrates low-cost context-copy proposals with shallow screening and continuous pipeline admission to achieve robust, scalable acceleration.

\paragraph{Self-speculation and intermediate review.}
Skipping layers, exiting early, or using quantized state can create drafts from the target model itself~\citep{DraftVerify24,Kangaroo24,SWIFT25,QuantSpec25}. 
LayerSkip trains early readouts and continues verification from cached early-layer states~\citep{LayerSkip24}. 
Cascade Speculative Drafting places cheap statistical proposals below intermediate reviewers, including its Max-Gram copier~\citep{Cascade24}. 
CAS-Spec constructs adaptive target-derived draft hierarchies that can include prompt lookup~\citep{CASSpec25}. 
HiSpec accumulates intermediate-checked prefixes before full verification~\citep{HiSpec25}; concurrent ECHO screens retrieval-based token trees in early layers and continues from their cached states~\citep{ECHO26}. 
While these approaches establish intermediate review and state continuation, \CL\ integrates hard prefix screening directly into shared-weight recurrent execution, allowing surviving blocks and their successors to advance concurrently. 
Building upon this design, our factorial analysis evaluates how screening effectiveness shifts when new successors are admitted prior to full-depth verification completion.

\section{Limitations} 
\label{sec:limitations}
Our evaluation has several limitations. 
First, due to resource constraints, our hardware evaluation is restricted to a single RTX 5090 GPU per worker, leaving validation across broader hardware setups (e.g., multi-GPU or alternative architectures) to future work. 
Second, execution under BF16 precision introduces numerical sensitivity: variations in batch shapes and execution schedules can alter intermediate token predictions and subsequent continuations, meaning that outputs are not strictly deterministic across all hardware configurations.

%% file: sections/conclusion.tex
\section{Conclusion}
\label{sec:conclusion}
In this work, we introduced \CL, a training-free decoding framework that accelerates looped Transformers by coupling low-cost context-copy proposals with early shallow screening and continuous pipeline scheduling. 
Our extensive evaluations across diverse Base and Thinking architectures demonstrate substantial speedups—ranging from $2.74\times$ to $10.70\times$. 
Beyond raw efficiency, our factorial analysis uncovers an important synergy: shallow screening is indispensable for making continuous speculative admission viable, preventing erroneous tails from flooding the pipeline. 
Furthermore, our hyperparameter studies demonstrate that \CL\ requires only lightweight calibration when transferring to new model architectures. 
Lastly, our analysis shows that \CL\ achieves greater speedups with rich matching context spans and sufficient response lengths to amortize startup overheads. 
We expect \CL\ will serve as a strong foundation for future research in efficient recurrent and test-time compute scaling. 

%% file: sections/ai_use.tex
\section{AI Use}
\label{sec:ai_use}

This work builds on a finding made by autonomous agents run by the authors on Station, a multi-agent research environment. Given the open-ended task of accelerating looped transformers, the agents discovered that copying past matching chunks and screening at a shallow position significantly accelerates looped transformers across models and benchmarks. The agent published its findings at \url{https://dualverse.ai/copyloop/4/}.

The authors considered this finding potentially useful to the broader research community and therefore reproduced and extended the experiments outside Station, using additional models and datasets. The authors also revised the manuscript to make the findings accessible to researchers.

%% file: sections/appendix.tex
\section{Decoding procedure and implementation}
\label{app:procedure}
This appendix provides pseudocode for the decoder in Section~\ref{sec:method} and an optional extension that branches at a second readout depth. We use $D$ for the full recurrent depth, as in the main method.

\subsection{Decoding algorithm}
Algorithm~\ref{alg:single} specifies initialization, continuous advancement, causal commitment, and recovery for the single-readout-depth decoder. 

\begin{algorithm}[htbp]
\caption{Single-readout-depth CopyLoop}
\label{alg:single}
\small
\begin{tabbing}
\quad\=\quad\=\quad\=\quad\=\kill
\textbf{Input:} prompt $x$; recurrent depth $D$; screening depth $1\le d_s<D$;\\
\qquad readout $g_d(p)$ at depth $d$ for prefix $p$; copied-tail limit $K$;\\
\qquad admission limit $C$ (token positions); output budget $L$; stop predicate $\mathrm{Stop}$\\
\textbf{Output:} accepted continuation $y$\\[2pt]

$h\gets\proc{Prefill}(x,D)$; $y\gets[h]$; $Q\gets\varnothing$\\
\textbf{if} $\mathrm{Stop}(y)$ or $|y|=L$ \textbf{then return} $y$\\
$Q\gets\proc{Admit}(Q,x,h)$\\
\textbf{while} $\neg\mathrm{Stop}(y)$ and $|y|<L$ \textbf{do}\\
\> $Q\gets\proc{Advance}(Q)$\\
\> \textbf{for each} block $B$ first reaching $d_s$ \textbf{do}\\
\>\> $m\gets$ longest copied prefix satisfying Eq.~\eqref{eq:screen}\\
\>\> retain the head and the first $m$ copied tokens of $B$\\
\>\> release the remaining suffix and its descendants\\
\>\> $p\gets$ prefix through the retained endpoint of $B$\\
\>\> $Q\gets\proc{Admit}(Q,p,g_{d_s}(p))$\\
\> \textbf{while} $Q$ contains a completed depth-$D$ readout for $s=x\cat y$ \textbf{do}\\
\>\> $t\gets$ the cached depth-$D$ prediction for $s$\\
\>\> $y\gets y\cat t$\\
\>\> \textbf{if} $\mathrm{Stop}(y)$ or $|y|=L$ \textbf{then return} $y$\\
\>\> discard branches inconsistent with $s\cat t$\\
\>\> \textbf{if} no retained successor has head $t$ after prefix $s$ \textbf{then}\\
\>\>\> $Q\gets\proc{Admit}(Q,s,t)$\\
\textbf{return} $y$\\[2pt]

\textbf{procedure} $\proc{Admit}(Q,p,h)$\\
\> construct a block with prefix $p$, head $h$, and at most $K$ copied tokens\\
\> \textbf{if} adding the block would make $\operatorname{size}(Q)>C$ \textbf{then}\\
\>\> remove its copied tail and retain only the head $h$\\
\> enqueue the block in $Q$; \textbf{return} $Q$
\end{tabbing}

Here $\cat$ denotes sequence concatenation, and $Q$ contains live
positions together with completed readouts. The prefill token $h$ is
already accepted and is therefore not emitted a second time.
\end{algorithm}

  \subsection{Branching at a Second Readout Depth}
  \label{app:two-checkpoint}
  The single-depth decoder can be extended to propose an alternative successor at a second readout depth $d_s<d_2<D$. Each block records the prediction $h_s$ at its retained endpoint at $d_s$. At $d_2$, the same endpoint produces $h_2$; an additional child is admitted only if $h_2\ne h_s$. This second readout adds a branch rather than
  repeating the copied-prefix screen. Each child obtains its own copied tail and is screened at $d_s$. Full-depth verification retains the continuation compatible with the verified prefix, removes incompatible descendants, and creates a correction block if no compatible continuation exists.

  We evaluate this extension on Ouro-1.4B Base and Thinking, comparing $d_s=1$ alone with an additional readout at $d_2=2$. For Base, we use 32 HumanEval+ and 32 GSM8K problems, with $K=6$ and a 1,024-token output budget. For Thinking, we use 32 HumanEval+ and 32 MATH-500 problems, with $K=2$, Thinking enabled, and an 8,192-token output
  budget. Within each model--task pair, both variants use identical prompts, stopping rules, and budgets. We perform two paired timing repeats with reversed configuration order, keeping each paired block on the same GPU. Aggregate decode throughput is the total number of decoded tokens divided by the total decode time, excluding prefill,
  first-token generation, startup, and warmup. AR normalization reuses previously measured records for the same problems; the single-depth and second-readout variants are measured contemporaneously.

  Table~\ref{tab:second-readout} shows that the benefit depends on the workload. On Base, the second readout improves decode throughput by $4.02\%$ on HumanEval+ and $11.66\%$ on GSM8K. On Thinking, the changes are $+1.79\%$ on HumanEval+ and $-0.47\%$ on MATH-500. The corresponding paired-task bootstrap 95\% confidence intervals for
  Thinking are $[-6.14\%,9.73\%]$ and $[-13.37\%,17.57\%]$, respectively, providing no clear evidence of a throughput improvement on these panels. These measurements use naturally generated BF16 outputs, whose lengths can differ between variants. The additional branches increase scheduling and state-management complexity, while the measured
  throughput gains are not consistent across the evaluated settings. We therefore retain the single-depth decoder as the main configuration.

\begin{table}[htbp]
\centering
\caption{\textbf{Second-readout branching on Ouro-1.4B.} Both variants use $d_s=1$ and the same copy length, output budget, and timing protocol within each model--task pair. Each pair contains 32 problems and two timing repeats. Speedups are aggregate decode-throughput ratios relative to AR for the same model and problems; gain is the
relative decode-throughput improvement over the single-depth variant, calculated from unrounded measurements.}
\label{tab:second-readout}
\small
\begin{dualversetable}
\begin{tabular}{llcccc}
\toprule
\DualverseHeadRow
\DualverseHead[l]{Model} & \DualverseHead[l]{Task} & \DualverseHead{$d_2$} & \DualverseHead{Single Depth\\($\times$ AR)} & \DualverseHead{Second Readout\\($\times$ AR)} & \DualverseHead{Gain (\%)} \\
\midrule

\rowcolor{DualverseB!30}
\multicolumn{6}{l}{\textbf{Base models}} \\
Ouro-1.4B Base & HumanEval+ & 2 & 5.86 & 6.09 & +4.02 \\
 & GSM8K & 2 & 3.36 & 3.75 & +11.66 \\

\rowcolor{DualverseB!30}
\multicolumn{6}{l}{\textbf{Thinking models}} \\
Ouro-1.4B Thinking & HumanEval+ & 2 & 3.36 & 3.42 & +1.79 \\
 & MATH-500 & 2 & 3.34 & 3.33 & -0.47 \\

\bottomrule
\end{tabular}
\end{dualversetable}
\end{table}

\section{Experimental Setup and Evaluation Protocols}
\label{app:experimental-protocols}
\label{app:protocol}
This section expands the experimental setup in Section~\ref{sec:experiments}. The main benchmark evaluation, screening/scheduling ablation, hyperparameter study, and workload diagnostics use distinct panels. Their sample sizes and output budgets are specified separately below.

\subsection{Hardware and Timing Metrics}
\label{app:measurement}
Experiments use BF16 weights on RTX 5090 GPUs with 32\,GiB memory and one active request per worker. Methods are compared under matched prompts, stopping rules, and output budgets within each model--task panel. Model loading, explicit warmup, and process startup are excluded from the reported timing.

The primary metric is aggregate decoding throughput, measured in tokens per second (TPS), computed as total generated tokens excluding the prefill-produced first token divided by their corresponding total decoding time. Speedup is the ratio of this quantity to the same model's native AR throughput, as defined in Eq.~\eqref{eq:speedup}. It is a ratio of aggregate throughputs, rather than an average of per-request speedups. Because BF16 execution schedules can yield different continuations, generated token totals can differ across methods.

The response-length analysis in Figure~\ref{fig:workloads}b instead uses request time, including prefill, decoding, and request handling. The ratio of summed request times and the ratio of decode TPS therefore answer different questions. 

\subsection{Models, Main Benchmark Panels, and Scoring}
\label{app:main-protocol}
The main evaluation includes four Base models and two Thinking models. Table~\ref{tab:model-settings} specifies the recurrent depths and CopyLoop defaults. Each CopyLoop configuration uses one screening depth, followed by final full-depth verification. Optional second-readout branching in Appendix~\ref{app:two-checkpoint} is separate from these main configurations.

\begin{table}[htbp]
\centering
\caption{Model configurations for the main evaluation. Depths count recurrent iterations. $K$ is the maximum copied-tail length.}
\label{tab:model-settings}
\begin{dualversetable}
\setlength{\tabcolsep}{3pt}
\renewcommand{\arraystretch}{1.04}
\setlength{\DualverseHeadPad}{2.6pt}
\begin{tabular}{>{\raggedright\arraybackslash}p{\dimexpr0.490\linewidth-2\tabcolsep\relax}>{\centering\arraybackslash}p{\dimexpr0.250\linewidth-2\tabcolsep\relax}>{\centering\arraybackslash}p{\dimexpr0.130\linewidth-2\tabcolsep\relax}>{\centering\arraybackslash}p{\dimexpr0.130\linewidth-2\tabcolsep\relax}}
\toprule
\DualverseHeadRow
\DualverseHead[l]{Model} & \DualverseHead{Full recurrent\\depth} & \DualverseHead{$d_s$} & \DualverseHead{$K$} \\
\rowcolor{DualverseB!30}
\multicolumn{4}{l}{\textbf{Base models}} \\
Ouro-1.4B / Ouro-2.6B & 4 & 1 & 6 \\
Raven-TinyLlama & 32 & 4 & 6 \\
Huginn-3.5B & 32 & 8 & 6 \\
\rowcolor{DualverseB!30}
\multicolumn{4}{l}{\textbf{Thinking models}} \\
Ouro-1.4B Thinking / Ouro-2.6B Thinking & 4 & 1 & 2 \\
\bottomrule
\end{tabular}
\end{dualversetable}
\end{table}

\begin{table}[htbp]
\centering
\caption{Main benchmark panels and output budgets, per model and method.}
\label{tab:protocol}
\begin{dualversetable}
\setlength{\tabcolsep}{3pt}
\renewcommand{\arraystretch}{1.04}
\setlength{\DualverseHeadPad}{2.6pt}
\begin{tabular}{>{\raggedright\arraybackslash}p{\dimexpr0.250\linewidth-2\tabcolsep\relax}>{\centering\arraybackslash}p{\dimexpr0.150\linewidth-2\tabcolsep\relax}>{\centering\arraybackslash}p{\dimexpr0.230\linewidth-2\tabcolsep\relax}>{\centering\arraybackslash}p{\dimexpr0.370\linewidth-2\tabcolsep\relax}}
\toprule
\DualverseHeadRow
\DualverseHead[l]{Benchmark} & \DualverseHead{Problems} & \DualverseHead{Output budget} & \DualverseHead{Primary score} \\
\rowcolor{DualverseB!30}
\multicolumn{4}{l}{\textbf{Base models}} \\
HumanEval+ & 164 & 1,024 & EvalPlus tests \\
MBPP+ & 378 & 1,024 & EvalPlus tests \\
GSM8K & 1,319 & 1,024 & Numerical answer \\
MATH-500 & 500 & 2,048 & Symbolic answer \\
\rowcolor{DualverseB!30}
\multicolumn{4}{l}{\textbf{Thinking models}} \\
AIME 2024 & 30 & 8,192 & Integer answer \\
AIME 2025 & 30 & 8,192 & Integer answer \\
BeyondAIME & 64 & 8,192 & Integer answer \\
MATH-500 & 64 & 8,192 & Symbolic answer \\
\bottomrule
\end{tabular}
\end{dualversetable}
\end{table}

\paragraph{Base prompts and evaluation.}
HumanEval uses code-completion prompts and MBPP+ uses code-generation prompts; both are scored with the extended EvalPlus tests~\citep{EvalPlus23}. GSM8K uses numerical answer extraction, and MATH-500 uses symbolic checking with \texttt{math-verify}. MATH-500 uses zero-shot prompting and end-of-text stopping. No beginning-of-sequence token is added in these Ouro Base configurations. 

\paragraph{Thinking prompts and final-answer scoring.}
\label{app:thinking-protocol}
The Thinking panel contains 188 problems per model: 30 each from AIME 2024 and AIME 2025, and 64 each from BeyondAIME and MATH-500. MATH-500 problems are selected proportionally across subject and difficulty strata. 
 Generation is greedy, with an 8,192-token budget and EOS stopping.

We score only the final-answer segment. Unfinished thinking, empty final answers, and malformed control-token sequences receive zero; budget-limited responses remain scoreable if a valid final answer is present. AIME requires an integer from 0 to 999, while BeyondAIME accepts integers without this range restriction. Answer extraction prioritizes the last boxed answer, followed by an explicit answer phrase and then a bare integer. MATH-500 uses \texttt{math-verify} on the final-answer segment.

\subsection{Baseline Implementations and Configuration}
\label{app:baselines}
Within each model--task panel, all methods use the same precision, problems, prompts, stopping rules, and output budgets. Our native implementation uses an SGLang-based recurrent engine~\citep{SGLang24}, recurrent kernels, and CUDA graphs.

The WaveFront port advances active positions each tick and verifies completed full-depth predictions, retaining the wave or restarting after a correction. It maintains depth-specific K/V states and follows the official rule of suppressing shallow proposals on rejection ticks. Cross-recurrence K/V sharing, sampling, and early exit are disabled.

\subsection{Hyperparameter Calibration and Transfer Protocol}
\label{app:calibration-protocol}
For each of the six models, we evaluate 20 calibration datasets and four evaluation benchmarks, using eight fixed examples per dataset. We test $K\in\{2,4,6,8\}$ with $d_s\in\{1,2\}$ for Ouro and $d_s\in\{4,8,16\}$ for Raven-TinyLlama and Huginn-3.5B. The two additional Raven models calibrate depth at fixed $K=6$ and use the full $(d_s,K)$ grid on the four Base evaluation benchmarks. Their evaluation grids predate calibration, making this a retrospective transfer analysis across distinct datasets.

\paragraph{Datasets and Fixed Examples.}
Table~\ref{tab:calibration-datasets} lists the 20 calibration datasets shared by all eight models, covering diverse tasks and excluding the Base and Thinking evaluation benchmarks.

\begin{table}[htbp]
\centering
\caption{Calibration datasets, domains, and splits.}
\label{tab:calibration-datasets}
\begin{dualversetable}
\setlength{\tabcolsep}{3pt}
\renewcommand{\arraystretch}{1.04}
\setlength{\DualverseHeadPad}{2.6pt}
\begin{tabular}{>{\raggedright\arraybackslash}p{\dimexpr0.350\linewidth-2\tabcolsep\relax}>{\raggedright\arraybackslash}p{\dimexpr0.450\linewidth-2\tabcolsep\relax}>{\raggedright\arraybackslash}p{\dimexpr0.200\linewidth-2\tabcolsep\relax}}
\toprule
\DualverseHeadRow
\DualverseHead[l]{Dataset} & \DualverseHead[l]{Domain} & \DualverseHead[l]{Split} \\
SVAMP & Mathematical reasoning & test \\
SQuAD & Extractive QA & validation \\
DialogSum & Dialogue summarization & test \\
E2E NLG & Data-to-text generation & test \\
CommonGen & Constrained generation & validation \\
OPUS-100 & Translation & test \\
BIG-Bench Hard & Reasoning and word sorting & test \\
MMLU-Pro & Knowledge and reasoning & test \\
DROP & Passage-based reasoning & validation \\
PubMedQA & Biomedical QA & train \\
WebNLG & RDF-to-text generation & test \\
WikiSQL & Text-to-SQL & test \\
CNN/DailyMail & News summarization & test \\
XSum & News summarization & test \\
HotpotQA & Multi-hop QA & validation \\
IFEval & Instruction following & train \\
WritingPrompts & Creative writing & test \\
WikiText-103 & Text continuation & test \\
Wikipedia & Text continuation & train \\
APPS & Introductory programming & test \\
\bottomrule
\end{tabular}
\end{dualversetable}
\end{table}

We select eight examples deterministically from frozen candidate panels, balancing categories or subtasks for calibration. All configurations within each model--dataset panel share the same examples, prompts, and output budgets. Context-length filtering can yield different selections across model families. The three Raven models share source examples, while Thinking calibration reuses the corresponding Base examples with the official Thinking chat template. Evaluation uses separate eight-example subsets from the main benchmarks; Base and Thinking use different MATH-500 subsets.

\paragraph{Inputs and Output Budgets.}
Base calibration uses task-specific completion prompts, while Thinking calibration uses the official chat template with Thinking enabled. Table~\ref{tab:calibration-budgets} lists the Base calibration output caps; Thinking uses 8,192 tokens throughout. All configurations within a model--dataset panel share the same prompts, budgets, and stopping rules.

Evaluation uses 1,024 tokens for Base coding and GSM8K, and 2,048 for Base MATH-500, except that Raven models use 1,024 for MATH-500. Thinking evaluation uses 8,192 tokens. EOS and task-specific stopping rules remain active.

\begin{table}[htbp]
\centering
\caption{Base calibration output caps in tokens. The Raven column applies to all three Raven models. Thinking calibration uses 8,192 tokens for every dataset.}
\label{tab:calibration-budgets}
\begin{dualversetable}
\setlength{\tabcolsep}{3pt}
\renewcommand{\arraystretch}{1.04}
\setlength{\DualverseHeadPad}{2.6pt}
\begin{tabular}{>{\raggedright\arraybackslash}p{\dimexpr0.290\linewidth-2\tabcolsep\relax}>{\centering\arraybackslash}p{\dimexpr0.190\linewidth-2\tabcolsep\relax}>{\centering\arraybackslash}p{\dimexpr0.190\linewidth-2\tabcolsep\relax}>{\centering\arraybackslash}p{\dimexpr0.170\linewidth-2\tabcolsep\relax}>{\centering\arraybackslash}p{\dimexpr0.160\linewidth-2\tabcolsep\relax}}
\toprule
\DualverseHeadRow
\DualverseHead[l]{Dataset} & \DualverseHead{Ouro-1.4B} & \DualverseHead{Ouro-2.6B} & \DualverseHead{Raven} & \DualverseHead{Huginn} \\
SVAMP & 1024 & 512 & 1024 & 512 \\
SQuAD & 128 & 128 & 128 & 128 \\
DialogSum & 256 & 256 & 512 & 256 \\
E2E NLG & 256 & 256 & 256 & 256 \\
CommonGen & 128 & 128 & 128 & 128 \\
OPUS-100 & 256 & 256 & 512 & 256 \\
BIG-Bench Hard & 1024 & 1024 & 1024 & 512 \\
MMLU-Pro & 1024 & 1024 & 1024 & 512 \\
DROP & 256 & 256 & 256 & 256 \\
PubMedQA & 1024 & 1024 & 512 & 512 \\
WebNLG & 256 & 256 & 256 & 256 \\
WikiSQL & 512 & 256 & 512 & 256 \\
CNN/DailyMail & 512 & 512 & 622 & 512 \\
XSum & 512 & 256 & 256 & 256 \\
HotpotQA & 256 & 256 & 288 & 256 \\
IFEval & 2048 & 1024 & 1729 & 1024 \\
WritingPrompts & 2048 & 2048 & 1908 & 1024 \\
WikiText-103 & 256 & 256 & 256 & 256 \\
Wikipedia & 256 & 256 & 256 & 256 \\
APPS & 2048 & 1024 & 1095 & 1024 \\
\bottomrule
\end{tabular}
\end{dualversetable}
\end{table}

\paragraph{Timing and Throughput.}
Runs use greedy decoding, with each example--configuration pair timed once after warmup. The eight examples per dataset are divided into four blocks of two; within each block, all configurations run on the same GPU in a deterministically shuffled order.

Decode time includes host scheduling, screening, and final verification, while excluding prefill, first-token computation, loading, and warmup. For model $m$, dataset $t$, and configuration $(d_s,K)$, aggregate decoding throughput is
\begin{equation}
 v_{m,t}(d_s,K)=
 \frac{\sum_{i=1}^{8}N_i^{\rm dec}}
      {\sum_{i=1}^{8}T_i^{\rm dec}},
 \label{eq:calibration-tps}
\end{equation}
where $N_i^{\rm dec}$ counts committed output tokens after the first token and $T_i^{\rm dec}$ is the corresponding decode time. Generated token counts may differ across configurations because their continuations can differ.

\paragraph{Model-Level Calibration and Deployment Defaults.}
For each initial model $m$, we select a configuration $c=(d_s,K)$ that maximizes the equal-weight geometric mean of throughput across the 20 calibration datasets:
\begin{equation}
 \widehat c_m=(\widehat d_m,\widehat K_m)
 =\arg\max_{c\in\mathcal C_m}
 \left(
 \prod_{t\in\mathcal T_{\rm cal}}
 v_{m,t}(c)
 \right)^{1/20},
 \label{eq:calibration}
\end{equation}
Ties prefer smaller $d_s$, then smaller $K$. The two additional Raven models apply the same criterion over depth at fixed $K=6$. Table~\ref{tab:calibrated-configs} reports the selected configurations.

For deployment, we retain the calibrated depth $\widehat d_m$ and adopt a shared copied-tail limit $K_0=6$ for Base and $K_0=2$ for Thinking.

\paragraph{Depth, Copied-Tail Limit, and Full-Grid Comparisons.}
We use one deficit definition throughout. For a candidate configuration $c$ and a reference configuration $r$, the task-wise throughput deficit is
\begin{equation}
 D_{m,t}(c\mathbin\Vert r)
 =100\left[
 1-\frac{v_{m,t}(c)}{v_{m,t}(r)}
 \right].
 \label{eq:task-deficit}
\end{equation}
For depth transfer, $c=(\widehat d_m,K_0)$ and $r$ is the configuration with the best tested depth at the same $K_0$. For copied-tail sensitivity, $c=(\widehat d_m,K_0)$ and $r$ is the configuration with the best tested $K$ at the same $\widehat d_m$. For full-grid transfer, $r$ is the best tested $(d_s,K)$ pair for that benchmark. Overall deficits use the geometric mean of the same candidate/reference throughput ratios:
\begin{equation}
 D_m^{\rm Overall}(c\mathbin\Vert r_{1:4})
 =100\left[
 1-\left(
 \prod_{t=1}^{4}
 \frac{v_{m,t}(c)}{v_{m,t}(r_t)}
 \right)^{1/4}
 \right].
 \label{eq:overall-deficit}
\end{equation}
Table~\ref{tab:hyperparameter-transfer} reports the full-grid deficits using the deployment configuration $c=(\widehat d_m,K_0)$ and the per-benchmark full-grid optima as references. All optima are restricted to the tested configurations on the same eight examples per benchmark.

\section{Supplementary Experimental Results}
\label{app:performance}
The following four subsections follow Sections~\ref{sec:overall}--\ref{sec:regimes}: main decoding performance, screening/scheduling ablation, hyperparameter selection, and workload boundaries.

\subsection{Decoding Performance and Task Outcomes}
\label{app:main-results}
\label{app:thinking}
Table~\ref{tab:overall} reports accuracy and decoding speedup for the four Base and two Thinking models. Different BF16 execution schedules can produce different continuations and output lengths.

Table~\ref{tab:main-run-totals} provides absolute decoding statistics. Decode token counts exclude the prefill-produced first token and discarded speculative candidates, but include returned EOS/stop tokens. Timing follows Appendix~\ref{app:measurement}.

All requested responses remain in the scoring denominator. For Thinking models, unfinished thinking and empty final answers receive zero, while budget-limited responses remain scoreable if a valid final answer is present, following Appendix~\ref{app:thinking-protocol}. The table reports output-budget hits and unfinished-thinking counts separately.

On MATH-500, Raven-TinyLlama uses a 1,024-token output budget because of its model-specific limit; the other Base models use 2,048 tokens.
\begin{table}[htbp]
\centering
\caption{Absolute decoding throughput in tokens per second (TPS). Bold indicates the highest throughput for each model--task pair. Thinking CopyLoop results use $K=2$.}
\label{tab:main-run-totals}
\begin{dualversetable}
\setlength{\tabcolsep}{3pt}
\renewcommand{\arraystretch}{1.04}
\setlength{\DualverseHeadPad}{2.6pt}
\begin{tabular}{>{\raggedright\arraybackslash}p{\dimexpr0.250\linewidth-2\tabcolsep\relax}>{\centering\arraybackslash}p{\dimexpr0.150\linewidth-2\tabcolsep\relax}>{\centering\arraybackslash}p{\dimexpr0.150\linewidth-2\tabcolsep\relax}>{\centering\arraybackslash}p{\dimexpr0.150\linewidth-2\tabcolsep\relax}>{\centering\arraybackslash}p{\dimexpr0.150\linewidth-2\tabcolsep\relax}>{\centering\arraybackslash}p{\dimexpr0.150\linewidth-2\tabcolsep\relax}}
\toprule
\DualverseHeadRow
\DualverseHead[l]{Task} & \DualverseHead{AR} & \DualverseHead{PLD} & \DualverseHead{LoopSpec} & \DualverseHead{WaveFront} & \DualverseHead{CopyLoop} \\
\rowcolor{DualverseB!30}
\multicolumn{6}{l}{\textbf{Ouro-1.4B --- Base}} \\
HumanEval+ & 103.03 & 281.41 & 333.42 & 332.06 & \textbf{622.32} \\
MBPP+ & 104.42 & 146.01 & 309.77 & 296.39 & \textbf{311.42} \\
GSM8K & 103.13 & 198.86 & 303.75 & 279.11 & \textbf{330.27} \\
MATH-500 & 99.14 & 184.62 & 295.45 & 273.37 & \textbf{335.09} \\
\rowcolor{DualverseB!30}
\multicolumn{6}{l}{\textbf{Ouro-2.6B --- Base}} \\
HumanEval+ & 53.89 & 137.31 & 182.06 & 183.10 & \textbf{340.43} \\
MBPP+ & 54.72 & 75.48 & \textbf{170.27} & 160.87 & 168.45 \\
GSM8K & 53.60 & 100.44 & 170.05 & 161.70 & \textbf{193.65} \\
MATH-500 & 50.36 & 93.39 & 159.24 & 152.56 & \textbf{195.30} \\
\rowcolor{DualverseB!30}
\multicolumn{6}{l}{\textbf{Raven-TinyLlama --- Base}} \\
HumanEval+ & 47.25 & 152.14 & 309.28 & 257.12 & \textbf{505.58} \\
MBPP+ & 47.38 & 105.11 & 283.62 & 226.19 & \textbf{356.94} \\
GSM8K & 47.14 & 104.15 & 257.25 & 185.94 & \textbf{316.42} \\
MATH-500 & 47.16 & 110.21 & 312.98 & 241.62 & \textbf{408.82} \\
\rowcolor{DualverseB!30}
\multicolumn{6}{l}{\textbf{Huginn-3.5B --- Base}} \\
HumanEval+ & 13.68 & 28.07 & 51.07 & 38.19 & \textbf{63.93} \\
MBPP+ & 13.70 & 19.46 & \textbf{44.55} & 35.45 & 42.50 \\
GSM8K & 13.69 & 21.85 & 42.72 & 35.04 & \textbf{46.10} \\
MATH-500 & 13.53 & 36.41 & 55.44 & 38.98 & \textbf{80.92} \\
\rowcolor{DualverseB!30}
\multicolumn{6}{l}{\textbf{Ouro-1.4B --- Thinking}} \\
AIME 2024 & 70.39 & 132.47 & 186.16 & 170.13 & \textbf{205.03} \\
AIME 2025 & 70.81 & 134.40 & 186.79 & 166.67 & \textbf{214.14} \\
BeyondAIME & 70.51 & 138.10 & 179.18 & 164.73 & \textbf{193.12} \\
MATH-500 & 80.23 & 150.88 & 224.19 & 202.69 & \textbf{258.81} \\
\rowcolor{DualverseB!30}
\multicolumn{6}{l}{\textbf{Ouro-2.6B --- Thinking}} \\
AIME 2024 & 36.60 & 72.12 & 102.47 & 94.27 & \textbf{119.69} \\
AIME 2025 & 36.23 & 69.86 & 98.10 & 90.68 & \textbf{113.57} \\
BeyondAIME & 36.16 & 71.02 & 96.52 & 88.38 & \textbf{114.86} \\
MATH-500 & 42.61 & 74.76 & 124.50 & 116.07 & \textbf{149.82} \\
\bottomrule
\end{tabular}
\end{dualversetable}
\end{table}

\subsubsection{BF16 Execution and Output Differences}
\label{app:numerics}

Different BF16 execution schedules can introduce minor numerical variations that alter greedy continuation paths. Specifically, small state fluctuations near an $\operatorname{argmax}$ tie can diverge a token and its subsequent generation history. The targeted diagnostics below illustrate this inherent numerical sensitivity rather than asserting a universal cause for every accuracy shift in Table~\ref{tab:overall}.

\paragraph{Internal State Sensitivity.}
Targeted inspections on shared prefixes reveal that hidden states can diverge across different execution schedules (e.g., at specific recurrent depth and block-layer locations), even when incoming key-value entries match exactly. Using an FP32 bridge starting from captured BF16 states further demonstrates that such internal fluctuations can occasionally shift the final $\operatorname{argmax}$ selection. However, these localized numerical divergences do not consistently propagate into systemic performance drops.

\paragraph{Impact on Task Accuracy.}
To evaluate whether output trajectory shifts translate to accuracy degradation, we conduct a controlled diagnostic panel on 64 GSM8K tasks. While altering attention-split settings modifies individual token trajectories (changing the specific set of divergent responses relative to native AR), the aggregate task accuracy remains entirely robust, with AR, dynamic CopyLoop, and fixed-split CopyLoop all achieving identical scores (59/64). These observations highlight a key distinction: execution-induced numerical sensitivity affects output text identity rather than task accuracy, confirming that minor benchmark variations reflect trajectory non-determinism rather than systematic algorithmic flaws.

\subsection{Screening and Scheduling Ablation}
\label{app:mechanism}
Table~\ref{tab:synergy-details} reports the complete numerical results for all eight panels in Figure~\ref{fig:synergy}.

\begin{table}[htbp]
\centering
\caption{Complete numerical results for Figure~\ref{fig:synergy}. Entries are decoding speedups relative to AR. C/S denote continuous/staged scheduling; +/-- denote shallow screening on/off.}
\label{tab:synergy-details}
\begin{dualversetable}
\setlength{\tabcolsep}{3pt}
\renewcommand{\arraystretch}{1.04}
\setlength{\DualverseHeadPad}{2.6pt}
\begin{tabular}{>{\raggedright\arraybackslash}p{\dimexpr0.280\linewidth-2\tabcolsep\relax}>{\centering\arraybackslash}p{\dimexpr0.180\linewidth-2\tabcolsep\relax}>{\centering\arraybackslash}p{\dimexpr0.180\linewidth-2\tabcolsep\relax}>{\centering\arraybackslash}p{\dimexpr0.180\linewidth-2\tabcolsep\relax}>{\centering\arraybackslash}p{\dimexpr0.180\linewidth-2\tabcolsep\relax}}
\toprule
\DualverseHeadRow
\DualverseHead[l]{Task} & \DualverseHead{C,+} & \DualverseHead{C,--} & \DualverseHead{S,+} & \DualverseHead{S,--} \\
\rowcolor{DualverseB!30}
\multicolumn{5}{l}{\textbf{Ouro-1.4B --- Base}} \\
HumanEval+ & \textbf{5.84} & 2.98 & 2.58 & 2.62 \\
GSM8K & \textbf{3.33} & 1.65 & 1.92 & 1.97 \\
\rowcolor{DualverseB!30}
\multicolumn{5}{l}{\textbf{Raven-TinyLlama --- Base}} \\
HumanEval+ & \textbf{9.71} & 2.83 & 2.79 & 2.78 \\
GSM8K & \textbf{6.23} & 1.41 & 1.88 & 1.87 \\
\rowcolor{DualverseB!30}
\multicolumn{5}{l}{\textbf{Ouro-1.4B --- Thinking}} \\
HumanEval+ & \textbf{3.31} & 1.77 & 1.69 & 1.67 \\
MATH-500 & \textbf{3.34} & 1.71 & 1.77 & 1.76 \\
\rowcolor{DualverseB!30}
\multicolumn{5}{l}{\textbf{Ouro-2.6B --- Thinking}} \\
HumanEval+ & \textbf{3.54} & 1.67 & 1.64 & 1.61 \\
MATH-500 & \textbf{3.41} & 1.67 & 1.65 & 1.67 \\
\bottomrule
\end{tabular}
\end{dualversetable}
\end{table}

\subsection{Hyperparameter Selection and Transfer}
\label{app:calibration}

To understand how our hyperparameter settings generalize, this subsection evaluates our approach across three dimensions: 
(1) the gap between calibration, joint optimization, and deployment configurations; 
(2) the transferability and sensitivity of the screening depth $d_s$ and copied-tail limit $K$; and 
(3) a diagnostic case study on output-trajectory sensitivity.

\subsubsection{Calibration, Joint Optimization, and Deployment Configurations}
Table~\ref{tab:calibrated-configs} compares calibration on 20 datasets with a single fixed configuration jointly optimized on the four main evaluation benchmarks. Both criteria use an equal-weight geometric mean of decode throughput over their respective datasets, following the protocol in Appendix~\ref{app:calibration-protocol}. Let $\widehat c_m=(\widehat d_m,\widehat K_m)$ denote the calibrated configuration and let $c_m^{\rm joint}$ denote the tested configuration that maximizes this mean over the four evaluation benchmarks. For a candidate configuration $c$ and a reference configuration $r$, we define the throughput deficit as
\begin{equation}
 D_m(c\mathbin\Vert r)
 =100\left[1-\left(
 \prod_{t=1}^{4}
 \frac{v_{m,t}(c)}{v_{m,t}(r)}
 \right)^{1/4}\right].
 \label{eq:throughput-deficit}
\end{equation}
The two selections coincide for five of the six initial models. The exception is Ouro-1.4B Thinking: calibration selects $(1,4)$, whereas joint optimization on the main benchmarks selects $(2,4)$, with $D_m(\widehat c_m\mathbin\Vert c_m^{\rm joint})=3.90\%$. Deployment retains the calibrated screening depth and uses a mode-specific default $K_0$: $K_0=6$ for Base models and $K_0=2$ for Thinking models; these deployment configurations are distinct from the full calibrated pairs.

\begin{table}[htbp]
\centering
\caption{\textbf{Calibration, joint optimization, and deployment configurations.} Every pair is $(d_s,K)$. Calibrated configurations maximize the equal-weight geometric mean of decode TPS over 20 calibration datasets; joint-optimal configurations maximize it over the four main evaluation benchmarks, using eight fixed examples per benchmark. The reported deficit uses Eq.~\eqref{eq:throughput-deficit} with the calibrated configuration as candidate and the joint-optimal configuration as reference. The two additional Raven models calibrate only depth at fixed $K=6$; their joint-optimal configurations and deficits are not reported here (---).}
\label{tab:calibrated-configs}
\begin{dualversetable}
\setlength{\tabcolsep}{3pt}
\renewcommand{\arraystretch}{1.04}
\setlength{\DualverseHeadPad}{2.6pt}
\begin{tabular}{>{\raggedright\arraybackslash}p{\dimexpr0.260\linewidth-2\tabcolsep\relax}>{\centering\arraybackslash}p{\dimexpr0.180\linewidth-2\tabcolsep\relax}>{\centering\arraybackslash}p{\dimexpr0.180\linewidth-2\tabcolsep\relax}>{\centering\arraybackslash}p{\dimexpr0.180\linewidth-2\tabcolsep\relax}>{\centering\arraybackslash}p{\dimexpr0.200\linewidth-2\tabcolsep\relax}}
\toprule
\DualverseHeadRow
\DualverseHead[l]{Model} & \DualverseHead{Calibrated\\$(d_s,K)$} & \DualverseHead{Joint-optimal\\$(d_s,K)$} & \DualverseHead{Deployment\\$(d_s,K)$} & \DualverseHead{Cal.-to-joint\\deficit (\%)} \\
\rowcolor{DualverseB!30}
\multicolumn{5}{l}{\textbf{Base models}} \\
Ouro-1.4B & $(1,4)$ & $(1,4)$ & $(1,6)$ & 0.00 \\
Ouro-2.6B & $(1,4)$ & $(1,4)$ & $(1,6)$ & 0.00 \\
Raven-TinyLlama & $(4,6)$ & $(4,6)$ & $(4,6)$ & 0.00 \\
Huginn-3.5B & $(8,8)$ & $(8,8)$ & $(8,6)$ & 0.00 \\
\rowcolor{DualverseB!30}
\multicolumn{5}{l}{\textbf{Thinking models}} \\
Ouro-1.4B Thinking & $(1,4)$ & $(2,4)$ & $(1,2)$ & 3.90 \\
Ouro-2.6B Thinking & $(1,4)$ & $(1,4)$ & $(1,2)$ & 0.00 \\
\rowcolor{DualverseB!30}
\multicolumn{5}{l}{\textbf{Additional validation on larger Base models}} \\
Raven-Llama-3.2 & $(4,6)$ & --- & $(4,6)$ & --- \\
Raven-OLMo-2 & $(4,6)$ & --- & $(4,6)$ & --- \\
\bottomrule
\end{tabular}
\end{dualversetable}
\end{table}

Moving from aggregate optimization to task-specific performance, Table~\ref{tab:calibration-task-transfer} compares the full calibrated pair $\widehat c_m$ with each evaluation benchmark's best tested pair $c^*_{m,t}$. Each task-wise deficit uses the single-benchmark form of Eq.~\eqref{eq:throughput-deficit}, and Overall uses its four-benchmark geometric-mean form with $c=\widehat c_m$ and $r=c^*_{m,t}$. The Overall deficits range from $2.58\%$ to $6.75\%$. This comparison uses the calibrated copied-tail limit $\widehat K_m$, whereas Table~\ref{tab:hyperparameter-transfer} evaluates the mode-specific deployment default $K_0$. The two sets of deficits therefore answer different questions.

\begin{table}[htbp]
\centering
\caption{\textbf{Transfer of full calibrated configurations to individual evaluation benchmarks.} Each task cell reports the benchmark's best tested $(d_s,K)$ pair / the throughput deficit (\%) of the calibrated pair in Table~\ref{tab:calibrated-configs}. Overall gives equal weight to the four task-wise throughput ratios via their geometric mean. All comparisons use eight fixed examples per benchmark. These are deficits of the full calibrated pairs, not of the deployment defaults.}
\label{tab:calibration-task-transfer}
\begin{dualversetable}
\setlength{\tabcolsep}{3pt}
\renewcommand{\arraystretch}{1.04}
\setlength{\DualverseHeadPad}{2.6pt}
\begin{subtable}{\linewidth}
\centering
\phantomcaption
\label{tab:calibration-task-transfer-base}
\begin{tabular}{>{\raggedright\arraybackslash}p{\dimexpr0.240\linewidth-2\tabcolsep\relax}>{\centering\arraybackslash}p{\dimexpr0.150\linewidth-2\tabcolsep\relax}>{\centering\arraybackslash}p{\dimexpr0.150\linewidth-2\tabcolsep\relax}>{\centering\arraybackslash}p{\dimexpr0.150\linewidth-2\tabcolsep\relax}>{\centering\arraybackslash}p{\dimexpr0.150\linewidth-2\tabcolsep\relax}>{\centering\arraybackslash}p{\dimexpr0.160\linewidth-2\tabcolsep\relax}}
\toprule
\DualverseHeadRow
\DualverseHead[l]{Model} & \DualverseHead{HumanEval+} & \DualverseHead{MBPP+} & \DualverseHead{GSM8K} & \DualverseHead{MATH-500} & \DualverseHead{Overall\\deficit (\%)} \\
\rowcolor{DualverseB!30}
\multicolumn{6}{l}{\textbf{Base models}} \\
Ouro-1.4B & $(1,8)$ / 3.65 & $(1,2)$ / 3.67 & $(1,2)$ / 0.66 & $(1,6)$ / 2.33 & \textbf{2.59} \\
Ouro-2.6B & $(1,8)$ / 3.69 & $(1,2)$ / 2.47 & $(1,2)$ / 5.14 & $(1,4)$ / 0.00 & \textbf{2.84} \\
Raven-TinyLlama & $(4,6)$ / 0.00 & $(4,6)$ / 0.00 & $(4,6)$ / 0.00 & $(4,8)$ / 13.38 & \textbf{3.53} \\
Huginn-3.5B & $(8,8)$ / 0.00 & $(8,4)$ / 3.85 & $(8,2)$ / 6.32 & $(8,8)$ / 0.00 & \textbf{2.58} \\
\bottomrule
\end{tabular}
\end{subtable}
\par\medskip
\begin{subtable}{\linewidth}
\centering
\phantomcaption
\label{tab:calibration-task-transfer-thinking}
\begin{tabular}{>{\raggedright\arraybackslash}p{\dimexpr0.240\linewidth-2\tabcolsep\relax}>{\centering\arraybackslash}p{\dimexpr0.150\linewidth-2\tabcolsep\relax}>{\centering\arraybackslash}p{\dimexpr0.150\linewidth-2\tabcolsep\relax}>{\centering\arraybackslash}p{\dimexpr0.150\linewidth-2\tabcolsep\relax}>{\centering\arraybackslash}p{\dimexpr0.150\linewidth-2\tabcolsep\relax}>{\centering\arraybackslash}p{\dimexpr0.160\linewidth-2\tabcolsep\relax}}
\toprule
\DualverseHeadRow
\DualverseHead[l]{Model} & \DualverseHead{AIME 2024} & \DualverseHead{AIME 2025} & \DualverseHead{BeyondAIME} & \DualverseHead{MATH-500} & \DualverseHead{Overall\\deficit (\%)} \\
\rowcolor{DualverseB!30}
\multicolumn{6}{l}{\textbf{Thinking models}} \\
Ouro-1.4B Thinking & $(1,4)$ / 0.00 & $(1,2)$ / 9.12 & $(2,4)$ / 6.67 & $(2,4)$ / 10.87 & \textbf{6.75} \\
Ouro-2.6B Thinking & $(1,4)$ / 0.00 & $(1,6)$ / 7.23 & $(1,6)$ / 4.01 & $(1,4)$ / 0.00 & \textbf{2.86} \\
\bottomrule
\end{tabular}
\end{subtable}
\end{dualversetable}
\end{table}

\subsubsection{Hyperparameter Transfer and Sensitivity ($d_s$ and $K$)}
We next analyze how individual hyperparameters transfer across tasks. 
First, regarding \textbf{screening-depth transfer}, at the mode-specific default $K$, the calibrated screening depth is optimal in 20 of the 24 initial model--benchmark panels and all eight panels for the two additional Raven models. Table~\ref{tab:depth-transfer-controlled} lists the four exceptions, whose throughput deficits range from $0.69\%$ to $6.81\%$.

\begin{table}[htbp]
\centering
\caption{Nonzero depth-only throughput deficits at the deployment configurations in Table~\ref{tab:calibrated-configs}. Each reference optimizes only $d_s$ at the same default $K$. The remaining 28 model--benchmark panels have zero deficit.}
\label{tab:depth-transfer-controlled}
\begin{dualversetable}
\setlength{\tabcolsep}{3pt}
\renewcommand{\arraystretch}{1.04}
\setlength{\DualverseHeadPad}{2.6pt}
\begin{tabular}{>{\raggedright\arraybackslash}p{\dimexpr0.660\linewidth-2\tabcolsep\relax}>{\centering\arraybackslash}p{\dimexpr0.340\linewidth-2\tabcolsep\relax}}
\toprule
\DualverseHeadRow
\DualverseHead[l]{Benchmark} & \DualverseHead{Deficit (\%)} \\
\rowcolor{DualverseB!30}
\multicolumn{2}{l}{\textbf{Huginn-3.5B}} \\
MATH-500 & 6.81 \\
\rowcolor{DualverseB!30}
\multicolumn{2}{l}{\textbf{Ouro-1.4B Thinking}} \\
AIME 2024 & 4.46 \\
BeyondAIME & 2.30 \\
MATH-500 & 0.69 \\
\bottomrule
\end{tabular}
\end{dualversetable}
\end{table}

Second, regarding \textbf{copied-tail limit sensitivity}, Table~\ref{tab:copy-limit-tps} reports throughput across copied-tail limits at each model's adopted screening depth. The best $K$ varies across benchmarks; for example, Ouro-1.4B prefers $K=8,2,2,6$ on its four benchmarks.

\begingroup
\small
\setlength{\tabcolsep}{4pt}
\begin{dualversetable}
\setlength{\tabcolsep}{3pt}
\renewcommand{\arraystretch}{1.04}
\setlength{\DualverseHeadPad}{2.6pt}
\setlength{\LTcapwidth}{\textwidth}
\begin{longtable}{>{\raggedright\arraybackslash}p{\dimexpr0.250\linewidth-2\tabcolsep\relax}>{\centering\arraybackslash}p{\dimexpr0.150\linewidth-2\tabcolsep\relax}>{\centering\arraybackslash}p{\dimexpr0.150\linewidth-2\tabcolsep\relax}>{\centering\arraybackslash}p{\dimexpr0.150\linewidth-2\tabcolsep\relax}>{\centering\arraybackslash}p{\dimexpr0.150\linewidth-2\tabcolsep\relax}>{\centering\arraybackslash}p{\dimexpr0.150\linewidth-2\tabcolsep\relax}}
\caption{Decode TPS at the screening depths in Table~\ref{tab:calibrated-configs}. Bold marks the best $K$ per panel. The final column is the default-$K$ throughput deficit (\%) against the best $K$ at that same depth, using Eq.~\eqref{eq:throughput-deficit}.}
\label{tab:copy-limit-tps}\\
\toprule
\DualverseHeadRow
\DualverseHead[l]{Benchmark} & \DualverseHead{$K=2$} & \DualverseHead{$K=4$} & \DualverseHead{$K=6$} & \DualverseHead{$K=8$} & \DualverseHead{Deficit (\%)} \\
\midrule
\endfirsthead
\multicolumn{6}{l}{\textit{Table \thetable\ continued}}\\
\toprule
\DualverseHeadRow
\DualverseHead[l]{Benchmark} & \DualverseHead{$K=2$} & \DualverseHead{$K=4$} & \DualverseHead{$K=6$} & \DualverseHead{$K=8$} & \DualverseHead{Deficit (\%)} \\
\midrule
\endhead
\midrule
\multicolumn{6}{r}{\textit{Continued on next page}}\\
\endfoot
\bottomrule
\endlastfoot

\rowcolor{DualverseB!30}
\multicolumn{6}{l}{\textbf{Ouro-1.4B --- Base}} \\*
HumanEval+ & 488.00 & 579.05 & 565.12 & \textbf{600.96} & 5.96 \\*
MBPP+ & \textbf{361.18} & 347.91 & 294.41 & 316.17 & 18.49 \\*
GSM8K & \textbf{361.68} & 359.29 & 331.07 & 340.49 & 8.46 \\*
MATH-500 & 345.19 & 338.84 & \textbf{346.93} & 323.83 & 0.00 \\
\rowcolor{DualverseB!30}
\multicolumn{6}{l}{\textbf{Ouro-2.6B --- Base}} \\*
HumanEval+ & 280.44 & 336.56 & 339.32 & \textbf{349.46} & 2.90 \\*
MBPP+ & \textbf{172.53} & 168.26 & 150.73 & 149.08 & 12.64 \\*
GSM8K & \textbf{221.12} & 209.75 & 211.03 & 211.36 & 4.56 \\*
MATH-500 & 203.30 & \textbf{215.07} & 214.48 & 202.69 & 0.27 \\
\rowcolor{DualverseB!30}
\multicolumn{6}{l}{\textbf{Raven-TinyLlama --- Base}} \\*
HumanEval+ & 368.82 & 393.03 & \textbf{493.82} & 480.06 & 0.00 \\*
MBPP+ & 382.44 & 449.15 & \textbf{497.09} & 446.41 & 0.00 \\*
GSM8K & 270.48 & 307.77 & \textbf{321.14} & 273.34 & 0.00 \\*
MATH-500 & 370.77 & 407.98 & 419.17 & \textbf{483.94} & 13.38 \\
\rowcolor{DualverseB!30}
\multicolumn{6}{l}{\textbf{Huginn-3.5B --- Base}} \\*
HumanEval+ & 57.17 & 59.92 & 60.72 & \textbf{64.34} & 5.63 \\*
MBPP+ & 38.85 & \textbf{41.80} & 40.61 & 40.19 & 2.86 \\*
GSM8K & \textbf{47.48} & 46.74 & 46.60 & 44.47 & 1.84 \\*
MATH-500 & 66.50 & 52.41 & 78.68 & \textbf{108.35} & 27.39 \\
\rowcolor{DualverseB!30}
\multicolumn{6}{l}{\textbf{Ouro-1.4B --- Thinking}} \\*
AIME 2024 & 204.64 & \textbf{229.76} & 212.16 & 193.30 & 10.93 \\*
AIME 2025 & \textbf{232.69} & 211.46 & 207.61 & 187.83 & 0.00 \\*
BeyondAIME & \textbf{192.94} & 192.51 & 171.50 & 176.58 & 0.00 \\*
MATH-500 & 235.45 & 220.06 & \textbf{244.36} & 198.51 & 3.65 \\
\rowcolor{DualverseB!30}
\multicolumn{6}{l}{\textbf{Ouro-2.6B --- Thinking}} \\*
AIME 2024 & 121.37 & \textbf{133.18} & 114.82 & 117.92 & 8.87 \\*
AIME 2025 & 124.28 & 118.05 & \textbf{127.24} & 110.23 & 2.33 \\*
BeyondAIME & 109.71 & 106.35 & \textbf{110.80} & 96.38 & 0.99 \\*
MATH-500 & 128.08 & \textbf{142.59} & 128.78 & 125.36 & 10.18 \\
\rowcolor{DualverseB!30}
\multicolumn{6}{l}{\textbf{Raven-Llama-3.2 --- Base (additional validation)}} \\*
HumanEval+ & 387.32 & 461.97 & \textbf{517.61} & 514.43 & 0.00 \\*
MBPP+ & 269.15 & \textbf{307.05} & 304.88 & 273.61 & 0.71 \\*
GSM8K & 223.96 & \textbf{249.00} & 238.02 & 234.79 & 4.41 \\*
MATH-500 & 363.21 & \textbf{482.82} & 465.83 & 479.83 & 3.52 \\
\rowcolor{DualverseB!30}
\multicolumn{6}{l}{\textbf{Raven-OLMo-2 --- Base (additional validation)}} \\*
HumanEval+ & 253.93 & 270.33 & 313.68 & \textbf{316.60} & 0.92 \\*
MBPP+ & 188.31 & \textbf{364.48} & 326.37 & 225.31 & 10.46 \\*
GSM8K & 175.11 & \textbf{186.67} & 179.73 & 171.36 & 3.72 \\*
MATH-500 & 192.74 & 200.75 & \textbf{217.98} & 197.50 & 0.00 \\
\end{longtable}
\end{dualversetable}
\endgroup

Finally, regarding **deployment defaults and broader model validation**, against the per-benchmark full-grid optima, the 24 initial panels have a mean deficit of $6.20\%$ and a median of $4.60\%$, with per-model Overall deficits of $3.53\%$–$10.09\%$. Raven-Llama-3.2 and Raven-OLMo-2 have maximum task-wise deficits of $4.41\%$ and $10.46\%$, and Overall deficits of $2.18\%$ and $3.86\%$, respectively. These comparisons allow both depth and $K$ to change, whereas the final column in Table~\ref{tab:copy-limit-tps} varies only $K$.

\subsubsection{Diagnostic Case Study: Output-Trajectory Sensitivity}
The Huginn MATH-500 results illustrate how aggregate throughput can be heavily influenced by output trajectory shifts. At $d_s=8$, increasing $K$ from $6$ to $8$ drastically alters the generated sequence lengths—growing the total decoded tokens from $4,187$ to $5,850$ while leaving total decode time nearly unchanged (from $53.22$ to $53.99$ seconds). Consequently, throughput surges from $78.68$ to $108.35$ TPS, resulting in a large apparent deficit of $27.39\%$ for the default configuration.

As detailed in Table~\ref{tab:trajectory-diagnostic}, this throughput discrepancy stems from trajectory-level sensitivity. At $K=8$, specific prompts (e.g., /392 and /51) hit the maximum output cap with lengthy repetitive passages, whereas other instances (e.g., /479) terminate significantly earlier.

\begin{table}[htbp]
\centering
\caption{Selected Huginn MATH-500 outputs at $d_s=8$. Decode counts exclude the first token; 2,047 decoded tokens reach the 2,048-token output cap.}
\label{tab:trajectory-diagnostic}
\begin{dualversetable}
\setlength{\tabcolsep}{3pt}
\renewcommand{\arraystretch}{1.04}
\setlength{\DualverseHeadPad}{2.6pt}
\begin{tabular}{>{\raggedright\arraybackslash}p{\dimexpr0.300\linewidth-2\tabcolsep\relax}>{\centering\arraybackslash}p{\dimexpr0.100\linewidth-2\tabcolsep\relax}>{\centering\arraybackslash}p{\dimexpr0.220\linewidth-2\tabcolsep\relax}>{\centering\arraybackslash}p{\dimexpr0.200\linewidth-2\tabcolsep\relax}>{\centering\arraybackslash}p{\dimexpr0.180\linewidth-2\tabcolsep\relax}}
\toprule
\DualverseHeadRow
\DualverseHead[l]{Example} & \DualverseHead{$K$} & \DualverseHead{Decode Tokens} & \DualverseHead{Time (s)} & \DualverseHead{TPS} \\
MATH500/392 & 6 & 193 & 3.20 & 60.37 \\
 & 8 & 2,047 & 10.61 & 192.92 \\
\midrule
MATH500/51 & 6 & 307 & 6.31 & 48.65 \\
 & 8 & 2,047 & 11.20 & 182.75 \\
\midrule
MATH500/479 & 6 & 2,047 & 17.18 & 119.17 \\
 & 8 & 340 & 6.17 & 55.15 \\
\bottomrule
\end{tabular}
\end{dualversetable}
\end{table}

A closer inspection reveals that this anomaly is driven by trajectory-level sensitivity. Specifically, two repetitive outputs at $K=8$ account for $69.98\%$ of the total decoded tokens and $40.40\%$ of the decode time. Excluding these two outlier examples from both configurations completely reverses their throughput ranking, yielding $84.35$ TPS for $K=6$ versus $54.57$ TPS for $K=8$. While this post-hoc check highlights how individual generation paths can skew aggregate metrics, all reported benchmark results strictly retain the full set of eight examples without selective filtering.

\subsection{Workload Boundaries: Copied Runs and Response Length}
\label{app:workloads}

\subsubsection{Saved-History Replay}
\label{app:replay}
At each saved output position, replay applies context lookup using suffix lengths four through one and the most recent match. Let $M$ denote the initial run of copied tokens matching the recorded continuation. With a tail limit of eight, Figure~\ref{fig:workloads}a reports the conditional survival probability
\begin{equation}
 \widehat H(j)=
 \frac{\sum_{i,p}\mathbf{1}\{M_{i,p}\geq j\}}
      {\sum_{i,p}\mathbf{1}\{M_{i,p}\geq1\}},
 \qquad 1\leq j\leq8,
 \label{eq:replay}
\end{equation}
where $i$ indexes requests and $p$ output positions. Positions are pooled across requests, so longer responses receive greater weight.

\begin{table}[htbp]
\centering
\caption{Saved-history replay with tail limit eight. The Matches column counts positions with $M\geq1$; $\widehat H(j)$ reports conditional survival probabilities.}
\label{tab:replay}
\begin{dualversetable}
\setlength{\tabcolsep}{3pt}
\renewcommand{\arraystretch}{1.04}
\setlength{\DualverseHeadPad}{2.6pt}
\begin{tabular}{>{\raggedright\arraybackslash}p{\dimexpr0.220\linewidth-2\tabcolsep\relax}>{\centering\arraybackslash}p{\dimexpr0.140\linewidth-2\tabcolsep\relax}>{\centering\arraybackslash}p{\dimexpr0.160\linewidth-2\tabcolsep\relax}>{\centering\arraybackslash}p{\dimexpr0.150\linewidth-2\tabcolsep\relax}>{\centering\arraybackslash}p{\dimexpr0.110\linewidth-2\tabcolsep\relax}>{\centering\arraybackslash}p{\dimexpr0.110\linewidth-2\tabcolsep\relax}>{\centering\arraybackslash}p{\dimexpr0.110\linewidth-2\tabcolsep\relax}}
\toprule
\DualverseHeadRow
\DualverseHead[l]{Task} & \DualverseHead{Requests} & \DualverseHead{Positions} & \DualverseHead{Matches} & \DualverseHead{$\widehat H(2)$} & \DualverseHead{$\widehat H(4)$} & \DualverseHead{$\widehat H(8)$} \\
\rowcolor{DualverseB!30}
\multicolumn{7}{l}{\textbf{Ouro-1.4B}} \\
HumanEval & 32 & 6,065 & 4,156 & .885 & .785 & .717 \\
GSM8K & 128 & 13,593 & 7,125 & .606 & .291 & .126 \\
\rowcolor{DualverseB!30}
\multicolumn{7}{l}{\textbf{Raven-TinyLlama}} \\
HumanEval & 32 & 6,185 & 4,312 & .858 & .721 & .595 \\
GSM8K & 128 & 13,060 & 6,765 & .600 & .271 & .089 \\
\bottomrule
\end{tabular}
\end{dualversetable}
\end{table}

Replay measures available copied runs along saved histories, rather than online acceptance rates or runtime savings.

\subsubsection{Response-Length Groups}
\label{app:distributions}
Figure~\ref{fig:workloads}b groups MBPP+ requests by LoopSpec's output length, retaining the same paired tasks for both methods. Within each bin, request speedup is the ratio of summed LoopSpec request time to summed CopyLoop request time. The figure's 95\% confidence intervals are obtained by resampling paired tasks within each bin; Table~\ref{tab:length-groups} reports counts and point estimates.

\begin{table}[htbp]
\centering
\caption{MBPP+ request speedup over LoopSpec, grouped by LoopSpec output length. Each model includes 378 problems.}
\label{tab:length-groups}
\begin{dualversetable}
\setlength{\tabcolsep}{3pt}
\renewcommand{\arraystretch}{1.04}
\setlength{\DualverseHeadPad}{2.6pt}
\begin{tabular}{>{\raggedright\arraybackslash}p{\dimexpr0.400\linewidth-2\tabcolsep\relax}>{\centering\arraybackslash}p{\dimexpr0.150\linewidth-2\tabcolsep\relax}>{\centering\arraybackslash}p{\dimexpr0.150\linewidth-2\tabcolsep\relax}>{\centering\arraybackslash}p{\dimexpr0.150\linewidth-2\tabcolsep\relax}>{\centering\arraybackslash}p{\dimexpr0.150\linewidth-2\tabcolsep\relax}}
\toprule
\DualverseHeadRow
\DualverseHead[l]{Metric} & \DualverseHead{1--32} & \DualverseHead{33--64} & \DualverseHead{65--128} & \DualverseHead{129+} \\
\rowcolor{DualverseB!30}
\multicolumn{5}{l}{\textbf{Ouro-1.4B}} \\
Problems & 149 & 111 & 85 & 33 \\
Request speedup & .981 & .991 & 1.029 & 1.302 \\
\rowcolor{DualverseB!30}
\multicolumn{5}{l}{\textbf{Ouro-2.6B}} \\
Problems & 159 & 121 & 75 & 23 \\
Request speedup & .912 & .946 & 1.010 & 1.429 \\
\bottomrule
\end{tabular}
\end{dualversetable}
\end{table}

Longer-response groups show greater speedups for both models, consistent with better amortization of fixed overheads. These groups use realized output lengths and do not establish a prompt-time routing rule or a universal length threshold.